\documentclass[]{article}

\usepackage[utf8]{inputenc}
\usepackage{amsmath,amssymb,amsthm}
\usepackage{microtype}
\usepackage{url}

\usepackage{listings}
\usepackage{multicol}
\usepackage{multirow}
\usepackage{hyperref}
\usepackage{caption}

\newcommand{\HT}{\ensuremath{\mathrm{HT}}}

\newcommand{\LTL}{\ensuremath{\mathrm{LTL}}}
\newcommand{\LTLf}{\ensuremath{\mathrm{LTL}_{\!f}}}

\newcommand{\LDL}{\ensuremath{\mathrm{LDL}}}
\newcommand{\LDLf}{\ensuremath{\mathrm{LDL}_{\!f}}}
\newcommand{\LDLo}{\ensuremath{\mathrm{LDL}_{\omega}}}
\newcommand{\DHT}{\ensuremath{\mathrm{DHT}}}
\newcommand{\DHTf}{\ensuremath{\mathrm{DHT}_{\!f}}}
\newcommand{\DHTo}{\ensuremath{\mathrm{DHT}_{\!\omega}}}
\newcommand{\DEL}{\ensuremath{\mathrm{DEL}}}
\newcommand{\DELf}{\ensuremath{{\DEL}_{\!f}}}
\newcommand{\DELo}{\ensuremath{{\DEL}_{\omega}}}

\newcommand{\DL}{\ensuremath{\mathrm{DL}}}
 \RequirePackage{bm}
\RequirePackage{textcomp}
\RequirePackage{upgreek}

\IfFileExists{outline.tex}{\input{outline}}{}

\newcommand{\alwaysF}{\ensuremath{\square}}

\newcommand{\eventuallyF}{\ensuremath{\Diamond}}

\IfFileExists{outline.tex}
        {\newcommand{\until}{\ensuremath{\mathbin{\mbox{\outline{$\bm{\mathsf{U}}$}}}}}}
        {\newcommand{\until}{\ensuremath{\mathbin{\bm{\mathsf{U}}}}}}
\IfFileExists{outline.tex}
        {\newcommand{\release}{\ensuremath{\mathbin{\mbox{\outline{$\bm{\mathsf{R}}$}}}}}}
        {\newcommand{\release}{\ensuremath{\mathbin{\bm{\mathsf{R}}}}}}
\IfFileExists{outline.tex}
        {}
        {}

\IfFileExists{outline.tex}
        {\newcommand{\finally}{\ensuremath{\mbox{\outline{$\bm{\mathsf{F}}$}}}}}
        {\newcommand{\finally}{\ensuremath{\bm{\mathsf{F}}}}}

\newcommand{\stp}{\ensuremath{\uptau}}

\newcommand{\dalways}[1]{\ensuremath{[#1]\,}}                        \newcommand{\deventually}[1]{\ensuremath{\langle#1\rangle\,}}          

\newcommand{\intervc}[2]{[#1..#2]}
\newcommand{\intervo}[2]{[#1..#2)}

\newcommand{\rangeo}[3]{#1 \in \intervo{#2}{#3}}

\newcommand{\tuple}[1]{\langle #1 \rangle}
\renewcommand{\H}{\ensuremath{\mathbf{H}}} \newcommand{\T}{\ensuremath{\mathbf{T}}}
\newcommand{\M}{\ensuremath{\mathbf{M}}}

 \newcommand{\sysfont}{\textit}

\newcommand{\Gringo}{\sysfont{Gringo}}
\newcommand{\abstem}{\sysfont{abstem}}

\newcommand{\asprilo}{\sysfont{asprilo}}

\newcommand{\clingo}{\sysfont{clingo}}

\newcommand{\gringo}{\sysfont{gringo}}

\newcommand{\stelp}{\sysfont{stelp}}

\newcommand{\telingo}{\sysfont{telingo}}

\newcommand{\aspif}{\sysfont{aspif}}

\providecommand{\Underscore}{\textunderscore}

\lstdefinelanguage{clingo}{basicstyle=\ttfamily,keywordstyle=[1]\bfseries,keywordstyle=[2]\bfseries,keywordstyle=[3]\bfseries,showstringspaces=false,literate={_}{\Underscore}1 {\%\%}{}0,escapeinside={\#(}{\#)},alsoletter={\#,\&},keywords=[1]{not,from,import,def,if,else,elif,return,while,break,and,or,for,in,del,and,class,with,as,is,yield,async},keywords=[2]{\#const,\#show,\#minimize,\#base,\#theory,\#count,\#external,\#program,\#script,\#end,\#heuristic,\#edge,\#project,\#show,\#sum},keywords=[3]{&,&dom,&sum,&diff,&show},morecomment=[l]{\#\ },morecomment=[l]{\%\ },morestring=[b]",stringstyle={\itshape},commentstyle={\color{darkgray}}}

\lstdefinelanguage{python}{basicstyle=\ttfamily,keywordstyle=[1]\bfseries,showstringspaces=false,literate={_}{\Underscore}{1},escapeinside={\#(}{\#)},alsoletter={\#,\&},keywords=[1]{not,from,import,def,if,else,elif,return,while,break,and,or,for,in,del,and,class,with,as,is,yield,async},morecomment=[l]{\#\ },morestring=[b]",stringstyle={\itshape},commentstyle={\color{darkgray}}}

\usepackage[normalem]{ulem}

\newtheorem{definition}{Definition}
\newtheorem{proposition}{Proposition}

\newtheorem{theorem}{Theorem}

\providecommand{\Next}{\text{\rm \raisebox{-.5pt}{\Large\textopenbullet}}}  \providecommand{\wNext}{\ensuremath{\widehat{\Next}}}

\newcommand{\eqdef}{\ensuremath{\mathbin{\raisebox{-1pt}[-3pt][0pt]{$\stackrel{\mathit{def}}{=}$}}}}
\newcommand{\fpbf}{\ensuremath{\bot}}
\newcommand{\tpbf}{\ensuremath{\top}}

\newcommand{\cl}[1]{\ensuremath{\mathit{cl}({#1})}}

\newcommand{\s}[1]{\ensuremath{q_{#1}\,}}

\newcommand{\last}{\ensuremath{\mathit{last}}}
\newcommand{\nnf}[1]{\ensuremath{\mathit{nnf}({#1})}}
\newcommand{\AFW}{\ensuremath{\mathrm{AFW}}}

\newcommand{\Rel}[2]{\parallel\!#1\!\parallel^{#2}}

\newcommand{\PV}{\ensuremath{\mathcal{P}}}
\newcommand{\myatom}{\ensuremath{a}}
\newcommand{\DBox}[1]{\dalways{#1}}
\newcommand{\DDia}[1]{\deventually{#1}}

\newcommand{\interp}{X}
\newcommand{\automata}{\ensuremath{\mathfrak{A}}}

\newcommand{\srtm}{\ensuremath{ST_m}}
\newcommand{\srtp}{\ensuremath{ST_p}}

\newcommand{\mso}{\text{mso}}

\newcommand{\runexample}{\ensuremath{\deventually{(\dalways{\stp^*} b)? \; ; \; \stp} a}}

\newcommand{\WF}[2]{\ensuremath{\mathfrak{#1}_{#2}}}
\newcommand{\WFT}{\WF{T}{\ignorespaces}}
\newcommand{\WFA}{\WF{A}{\ignorespaces}}
\newcommand{\WFM}{\WF{M}{\ignorespaces}}
\newcommand{\WFMm}{\WF{M}{\mathit{m}}}
\newcommand{\WFMs}{\WF{M}{\mathit{s}}}
\newcommand{\WFNC}{NC}

\newcommand{\fovar}[1]{\ensuremath{\mathbf{#1}}}

\usepackage{tikz} \usetikzlibrary{patterns,shapes.arrows,arrows,automata,positioning,trees,calc,shadows,positioning}
\usetikzlibrary{fit}

\begin{document}

\definecolor{stylered}{HTML}{F5D5CB}
\definecolor{stylegreen}{HTML}{D7ECD9}
\definecolor{styleyellow}{HTML}{F6F6EB}
\definecolor{stylepurple}{HTML}{D5D6EA}

\title{Automata for dynamic answer set solving: Preliminary report}
\author{
  Pedro Cabalar\\University of Corunna, Spain \and
  Mart\'{i}n Di\'eguez\\Universit\'e d'Angers, France \and
  Susana Hahn and Torsten Schaub\\University of Potsdam, Germany}
\date{}
\maketitle

\begin{abstract}
  We explore different ways of implementing temporal constraints expressed in an extension of
  Answer Set Programming (ASP) with language constructs from dynamic logic.
  Foremost, we investigate how automata can be used for enforcing such constraints.
  The idea is to transform a dynamic constraint into an automaton expressed in terms of a logic program that enforces
  the satisfaction of the original constraint.
  What makes this approach attractive is its independence of time stamps and the potential to detect unsatisfiability.
  On the one hand, we elaborate upon a transformation of dynamic formulas into alternating automata that relies on
  meta-programming in ASP.
  This is the first application of reification applied to theory expressions in \gringo.
  On the other hand, we propose two transformations of dynamic formulas into monadic second-order formulas.
  These can then be used by off-the-shelf tools to construct the corresponding automata.
  We contrast both approaches empirically with the one of the temporal ASP solver \telingo\ that directly maps dynamic
  constraints to logic programs.
  Since this preliminary study is restricted to dynamic formulas in integrity constraints, its
  implementations and (empirical) results readily apply to conventional linear dynamic logic, too.
\end{abstract}

 \section{Introduction}\label{sec:introduction}

Answer Set Programming (ASP~\cite{lifschitz99b}) has become a popular approach to solving knowledge-intense combinatorial search
problems due to its performant solving engines and expressive modeling language.
However, both are mainly geared towards static domains and lack native support for handling dynamic applications.
Rather \emph{change} is accommodated by producing copies of variables, one for each state.
This does not only produce redundancy but also leaves the ASP machinery largely uninformed about the temporal
structure of the problem.

This preliminary work explores alternative ways of implementing temporal (integrity) constraints in (linear) \emph{Dynamic Equilibrium Logic} (\DEL;~\cite{bocadisc18a,cadisc19a}) by using automata~\cite{hopull79a}.
On the one hand, \DEL\ is expressive enough to subsume more basic systems,
like (linear) Temporal Equilibrium Logic~\cite{agcadipevi13a,cakascsc18a} or even its metric variant~\cite{cadiscsc20a}.
On the other hand, our restriction to integrity constraints allows us to draw on work in conventional linear dynamic and
temporal logic (cf.\ Proposition~\ref{pro:dht:ldl}).
Although this amounts to using dynamic formulas to filter ``stable temporal models'' rather than
to let them take part in the formation of such models,
it allows us to investigate a larger spectrum of alternatives in a simpler setting.
Once fully elaborated, we plan to generalize our approach to the full setting.
Moreover, we are interested in implementing our approach by means of existing ASP systems,
which motivates our restriction to the finite trace variant of \DEL, called \DELf.

In more detail,
Section~\ref{sec:del} to~\ref{sec:ldlf2afw} lay the basic foundations of our approach
by introducing \DEL, some automata theory, and a translation from dynamic formula into alternating automata.
We then develop and empirically evaluate three different approaches.
First, the one based on alternating automata from Section~\ref{sec:ldlf2afw}.
This approach is implemented entirely in ASP and relies on meta-programming.
As such it is the first application of \gringo's reification machinery to user defined language constructs
(defined by a theory grammar; cf.\ \cite{kascwa17a}).
Second, the one elaborated in Section~\ref{sec:ldlf2mso}, proposing two alternative
transformations of dynamic formula into monadic second order formulas.
These formulas can then be passed to the off-the-shelf automata construction tool MONA~\cite{hejejoklparasa95a}
that turns them into deterministic automata.
And finally, the approach of \telingo~\cite{cakamosc19a,cadilasc20a},
transforming each dynamic constraint directly into a logic program.
All three approaches result in a program that allows us to sift out ``stable temporal models'' satisfying the original
dynamic constraints.
Usually, these models are generated by another logic program, like a planning encoding and instance.

\section{Linear Dynamic Equilibrium Logic}\label{sec:del} 

Given a set \PV\ of propositional variables (called \emph{alphabet}),
\emph{dynamic formulas} $\varphi$ and \emph{path expressions} $\rho$ are mutually defined by the pair of grammar rules:
\[
  \varphi  ::= \myatom \mid \bot \mid \top \mid \; \DBox{\rho} \varphi \; \mid \; \DDia{\rho} \varphi \qquad\qquad
  \rho     ::=  \stp \mid \varphi ? \mid \rho + \rho \mid \rho\mathrel{;}\rho \mid \rho^{\ast} . \]
This syntax is similar to the one of Dynamic Logic (\DL;~\cite{hatiko00a})
but differs in the construction of atomic path expressions:
While \DL\ uses a separate alphabet for \emph{atomic actions},
\LDL\ has a single alphabet \PV\ and
the only atomic path expression is the (transition) constant $\stp \not\in \PV$ (read as ``step'').
Thus, each $\rho$ is a regular expression formed with the constant $\stp$ plus the test construct $\varphi?$
that may refer to propositional atoms in the (single) alphabet \PV.
As with \LDL~\cite{giavar13a},
we sometimes use a propositional formula $\phi$ as a path expression and let it stand for $(\phi?;\stp)$.
This means that the reading of $\top$ as a path expression amounts to $(\top?;\stp)$
which is just equivalent to $\stp$, as we see below.
Another abbreviation is the sequence of $n$ repetitions of some expression $\rho$ defined as $\rho^0 \eqdef \top?$ and $\rho^{n+1} \eqdef \rho; \rho^n$.

The above language allows us to capture several derived operators, like the Boolean and temporal ones~\cite{cadisc19a}:
\[
  \arraycolsep=2pt\def\arraystretch{1.3}
  \begin{array}{rclp{10pt}rclp{10pt}rcl}
    \varphi \wedge \psi & \eqdef & \DDia{\varphi?} \psi          && \varphi \vee \psi & \eqdef & \DDia{\varphi?+\psi?} \top
    \\
    \varphi \to \psi & \eqdef & \DBox{\varphi?} \psi             && \neg \varphi & \eqdef & \varphi \to \bot
    \\
\Next \varphi & \eqdef & \DDia{\stp} \varphi                 && \wNext \varphi & \eqdef & \DBox{\stp} \varphi &\finally & \eqdef & \DBox{\stp}\bot
    \\
    \eventuallyF \varphi & \eqdef & \DDia{\stp^*} \varphi        && \alwaysF \varphi & \eqdef & \DBox{\stp^*} \varphi
    \\
    \varphi \until \psi & \eqdef & \DDia{(\varphi?;\stp)^*} \psi && \varphi \release \psi & \eqdef & (\psi \until (\varphi \wedge \psi)) \vee \alwaysF \psi
\end{array}
\]
All connectives are defined in terms of the dynamic operators \DDia{\cdot} and \DBox{\cdot}.
This involves the Booleans $\wedge$, $\vee$, and $\to$,
among which the definition of $\to$ is most noteworthy since it hints at the implicative nature of \DBox{\cdot}.
Negation $\neg$ is then expressed via implication, as usual in \HT.
Then, \DDia{\cdot} and \DBox{\cdot} also allow for defining the future temporal operators
\finally,
\Next,
\wNext,
\eventuallyF,
\alwaysF,
\until,
\release,
standing for
\emph{final,
next,
weak next,
eventually,
always,
until,} and
\emph{release}.
A  formula is \emph{propositional}, if all its connectives are Boolean,
and \emph{temporal}, if it includes only Boolean and temporal ones.
As usual, a \emph{(dynamic) theory} is a set of (dynamic) formulas.

For the semantics, we let $\intervc{a}{b}$ stand for the set $\{i \in \mathbb{N} \mid a \leq i \leq b\}$ and
$\intervo{a}{b}$ for $\{i \in \mathbb{N} \mid a \leq i < b\}$
for $a \in \mathbb{N}$ and $b \in \mathbb{N} \cup \{\omega\}$.
A \emph{trace} of length $\lambda$ over alphabet \PV\ is then defined as
a sequence $(H_i)_{\rangeo{i}{0}{\lambda}}$ of sets $H_i\subseteq\PV$.
A trace is \emph{infinite} if $\lambda=\omega$ and \emph{finite} otherwise, that is, $\lambda=n$ for some natural number $n \in \mathbb{N}$.
Given traces $\H=(H_i)_{\rangeo{i}{0}{\lambda}}$ and $\H'=(H'_i)_{\rangeo{i}{0}{\lambda}}$ both of length $\lambda$,
we write $\H\leq\mathbf\H'$ if $H_i\subseteq H'_i$ for each $\rangeo{i}{0}{\lambda}$;
accordingly, $\mathbf{H}<\mathbf{H'}$ iff both $\mathbf{H}\leq\mathbf{H'}$ and $\mathbf{H}\neq\mathbf{H'}$.

Although \DHT\ shares the same syntax as \LDL, its semantics relies on traces whose states are pairs of sets of atoms.
An \HT-trace is a sequence of pairs
\(
(\tuple{H_i,T_i})_{\rangeo{i}{0}{\lambda}}
\)
such that $H_i\subseteq T_i\subseteq \PV$ for any $\rangeo{i}{0}{\lambda}$.
As before, an \HT-trace is infinite if $\lambda=\omega$ and finite otherwise.
The intuition of using these two sets stems from \HT:
Atoms in $H_i$ are those that can be proved;
atoms not in $T_i$ are those for which there is no proof;
and, finally, atoms in $T_i\setminus H_i$ are assumed to hold, but have not been proved.
We often represent an \HT-trace as a pair of traces $\tuple{\H,\T}$ of length $\lambda$
where $\H=(H_i)_{\rangeo{i}{0}{\lambda}}$ and $\T=(T_i)_{\rangeo{i}{0}{\lambda}}$ such that $\H \leq \T$.
The particular type of \HT-traces that satisfy $\H=\T$ are called \emph{total}.

The overall definition of \DHT\ satisfaction relies on a double induction.
Given any \HT-trace $\M=\tuple{\H,\T}$,
we define \DHT\ satisfaction of formulas, namely, $\M,k \models \varphi$,
in terms of an accessibility relation for path expressions $\Rel{\rho}{\M} \subseteq \mathbb{N}^2$
whose extent depends again on $\models$ by double, structural induction.
\begin{definition}[\DHT\ satisfaction; \cite{cadisc19a}]\label{def:dht:satisfaction}
  An \HT-trace $\M=\tuple{\H,\T}$ of length $\lambda$ over alphabet \PV{}
  \emph{satisfies} a dynamic formula $\varphi$ at time point $\rangeo{k}{0}{\lambda}$,
  written \mbox{$\M,k \models \varphi$}, if the following conditions hold:
  \begin{enumerate}
  \item $\M,k \models \top$ and  $\M,k \not\models \bot$
  \item $\M,k \models \myatom$ if $\myatom \in H_k$ for any atom $\myatom \in \PV$
  \item \label{def:dhtsat.2} $\M, k \models \DDia{\rho} \varphi$
    if $\M,i \models \varphi$
    for some $i$ with $(k,i) \in \Rel{\rho}{\M}$
  \item \label{def:dhtsat.3} $\M, k \models \DBox{\rho} \varphi$
    if $\M',i \models \varphi$
    for all $i$ with $(k,i) \in \Rel{\rho}{\M'}$ \\ for both $\M'=\M$ and $\M'=\tuple{\T,\T}$
  \end{enumerate}
  where, for any \HT-trace $\M$, $\Rel{\rho}{\M} \subseteq \mathbb{N}^2$ is a relation on pairs of time points inductively defined as follows.
  \begin{enumerate}
  \setcounter{enumi}{4}
  \item $\Rel{\stp}{\M}            \ \eqdef\ \{ (k,k+1) \ \mid \rangeo{k,k+1}{0}{\lambda} \}$
  \item $\Rel{\varphi?}{\M}        \ \eqdef\ \{ (k,k) \mid  \M,k \models \varphi \}$
  \item $\Rel{\rho_1\mathrel{+}\rho_2}{\M} \ \eqdef\ \Rel{\rho_2}{\M} \cup \Rel{\rho_2}{\M}$
  \item
    \(
    \Rel{\rho_1\mathrel{;}\rho_2}{\M}  \ \eqdef\ \{ (k,i)  \mid {(k,j) \in \Rel{\rho_1}{\M}}\;\text{and }
    {(j,i) \in \Rel{\rho_2}{\M}}\;\text{for some } j \}
    \)
  \item $\Rel{\rho^*}{\M}          \ \eqdef\  \bigcup_{n\geq 0} \Rel{\rho^n}{\M}$
  \end{enumerate}
\end{definition}
An \HT-trace $\M$ is a \emph{model} of a dynamic theory $\Gamma$ if $\M,0 \models \varphi$ for all $\varphi \in \Gamma$.
We write $\DHT(\Gamma,\lambda)$ to stand for the set of \DHT\ models of length $\lambda$ of a theory $\Gamma$,
and define $\DHT(\Gamma) \eqdef \bigcup_{\lambda=0}^\omega \DHT(\Gamma,\lambda)$, that is, the whole set of models of $\Gamma$ of any length.
A formula $\varphi$ is a \emph{tautology} (or is \emph{valid}), written $\models \varphi$,
iff $\M,k \models \varphi$ for any \HT-trace \M\ and any $\rangeo{k}{0}{\lambda}$.
The logic induced by the set of all tautologies is called \emph{(Linear) Dynamic logic of Here-and-There} (\DHT\ for short).
We distinguish the variants \DHTo\ and \DHTf\ by restricting \DHT\ to infinite or finite traces, respectively.

\begin{proposition}\label{prop:ordering}
	For any $(x,y) \in \mathbb{N} \times \mathbb{N}$, path expression $\rho$ and trace $\M$, we have $(x,y) \in \Rel{\rho}{\M}$ implies $x \le y$.
\end{proposition}

\begin{proposition}[\cite{cadilasc20a,hakoti01a}]\label{prop:validities}
  The following formulas are \DHTf-valid.
  \begin{multicols}{2}
  \begin{enumerate}
  \item $\dalways{\rho_1 + \rho_2} \varphi \leftrightarrow\left(\dalways{\rho_1} \varphi \wedge \dalways{\rho_2} \varphi \right)$
  \item $\deventually{\rho_1 + \rho_2} \varphi \leftrightarrow\left(\deventually{\rho_1} \varphi \vee \deventually{\rho_2} \varphi \right)$
  \item $\dalways{\rho_1 ; \rho_2} \varphi \leftrightarrow\left(\dalways{\rho_1} \dalways{\rho_2} \varphi \right)$
  \item $\deventually{\rho_1 ; \rho_2} \varphi \leftrightarrow\left(\deventually{\rho_1}\deventually{\rho_2} \varphi \right)$
  \item $\dalways{\rho^*} \varphi \leftrightarrow\left( \varphi \wedge \dalways{\rho} \dalways{\rho^*} \varphi \right)$
  \item $\deventually{\rho^*} \varphi \leftrightarrow\left(\varphi \vee \deventually{\rho}\deventually{\rho^*} \varphi \right)$
  \end{enumerate}
  \end{multicols}
  \end{proposition}

We refrain from giving the semantics of \LDL~\cite{giavar13a}, since it corresponds to \DHT\ on total traces $\tuple{\T,\T}$~\cite{cadisc19a}.
Letting $\T,k \models \varphi$ denote the satisfaction of $\varphi$ by a trace \T\ at point $k$ in \LDL,
we have $\tuple{\T,\T},k \models \varphi$ iff $\T,k \models \varphi$ for $\rangeo{k}{0}{\lambda}$.
Accordingly, any total \HT-trace $\tuple{\T,\T}$ can be seen as the \LDL-trace \T.
As above, we denote infinite and finite trace variants as \LDLo\ and \LDLf, respectively.

The work presented in the sequel takes advantage of the following result
that allows us to treat dynamic formulas in occurring in integrity constraints as in \LDL:
\begin{proposition}\label{pro:dht:ldl}
  For any \HT-trace $\tuple{\H,\T}$ of length $\lambda$ and any dynamic formula $\varphi$, we have

  $\tuple{\H,\T}, k \models \neg\neg\varphi$ iff $\T,k \models \varphi$, for all $\rangeo{k}{0}{\lambda}$.
\end{proposition}

We now introduce non-monotonicity by selecting a particular set of traces called \emph{temporal equilibrium models}~\cite{cadisc19a}.
First, given an arbitrary set $\mathfrak{S}$ of \HT-traces, we define the ones in equilibrium as follows.
A total \HT-trace $\tuple{\T,\T} \in\mathfrak{S}$ is an \emph{equilibrium model} of $\mathfrak{S}$ iff
there is no other $\tuple{\H,\T} \in\mathfrak{S}$ such that $\H < \T$.
If this is the case, we also say that trace \T\ is a \emph{stable model} of $\mathfrak{S}$.
We further talk about \emph{temporal equilibrium} or \emph{temporal stable models} of a theory $\Gamma$ when $\mathfrak{S}=\DHT(\Gamma)$.
We write $\DEL(\Gamma,\lambda)$ and $\DEL(\Gamma)$ to stand for the temporal equilibrium models of $\DHT(\Gamma,\lambda)$ and $\DHT(\Gamma)$ respectively.
Note that stable models in $\DEL(\Gamma)$ are also \LDL-models of $\Gamma$.
Besides, as the ordering relation among traces is only defined for a fixed $\lambda$,
the set of temporal equilibrium models of $\Gamma$ can be partitioned by the trace length $\lambda$, that is,
$\bigcup_{\lambda=0}^\omega \DEL(\Gamma,\lambda) = \DEL(\Gamma)$.

(Linear) \emph{Dynamic Equilibrium Logic} (\DEL;~\cite{bocadisc18a,cadisc19a}) is the non-monotonic logic induced by temporal equilibrium models of dynamic theories.
We obtain the variants \DELo\ and \DELf\ by applying the corresponding restriction to infinite or finite traces, respectively.

As a consequence of Proposition~\ref{pro:dht:ldl}, the addition of formula $\neg \neg \varphi$ to a theory $\Gamma$ enforces that every temporal stable model of $\Gamma$ satisfies $\varphi$.
With this, we confine ourselves in Section~\ref{sec:ldlf2afw} and~\ref{sec:ldlf2mso} to \LDLf\ rather than \DELf.

In what follows, we consider finite traces only.
 \section{Automata}\label{sec:automata}

A \emph{Nondeterministic Finite Automaton} (NFA;~\cite{hopull79a}) is a tuple
\(
(\Sigma,Q,Q_0,\delta,F)
\),
where
$\Sigma$ is a finite nonempty alphabet,
$Q$ is a finite nonempty set of states,
$Q_0\subseteq Q$ is a set of initial states,
$\delta : Q \times \Sigma \to 2^Q$ is a transition function and
$F\subseteq Q$ a finite set of final states.
A run of an NFA $(\Sigma,Q,Q_0,\delta,F)$ on a word $a_0\cdots a_{n-1}$ of length $n$ for $a_i\in\Sigma$
is a finite sequence $q_0,\cdots,q_n$ of states
such that $q_0\in Q_0$
and $q_{i+1} \in \delta(q_i,a_i)$ for $0\leq i<n$.
A run is accepting if $q_n\in F$.
Using the structure of a NFA,
we can also represent a \emph{Deterministic Finite Automata} (DFA),
where
$Q_0$ contains a single initial state and
$\delta$ is restricted to return a single successor state.
A finite word $w\in\Sigma^*$ is accepted by an NFA, if there is an accepting run on $w$.
The language recognized by a NFA $\automata$ is defined as
\(
\mathcal{L}(\automata)=\{w \in \Sigma^* \mid \automata\text{ accepts }w\}
\).

An \emph{Alternating Automaton over Finite Words} (\AFW;~\cite{chkost81a,giavar13a}) is a tuple
\(
(\Sigma,Q,q_0,\delta,F)
\),
where
$\Sigma$ and $Q$ are as with NFAs,
$q_0$ is the initial state,
$\delta : Q \times \Sigma \to B^+(Q)$ is a transition function,
where $B^+(Q)$ stands for all propositional formulas built from $Q$, $\wedge$, $\vee$, $\tpbf$ and $\fpbf$,
and $F\subseteq Q$ is a finite set of final states.

A run of an \AFW\ $(\Sigma,Q,q_0,\delta,F)$ on a word $a_0\cdots a_{n-1}$ of length $n$ for $a_i\in\Sigma$,
is a finite tree $T$ labeled by states in $S$
such that
\begin{enumerate}
\item the root of $T$ is labeled by $q_0$,
\item if node $o$ at level $i$ is labeled by a state $q\in Q$ and $\delta(q,a_i) = \varphi$, then
  either $\varphi = \top$ or
$P\models\varphi$ for some $P \subseteq Q$ and $o$ has a child for each element in $P$,
\item the run is accepting if all leaves at depth $n$ are labeled by states in $F$.
\end{enumerate}
A finite word $w\in\Sigma^*$ is accepted by an \AFW, if there is an accepting run on $w$.
The language recognized by an \AFW\ $\automata$ is defined as
\(
\mathcal{L}(\automata)=\{w \in \Sigma^* \mid \automata\text{ accepts }w\}
\).

\AFW{s} can be seen as an extension of NFAs by universal transitions.
That is, when looking at formulas in $B^+(Q)$,
disjunctions represent alternative transitions as in NFAs, while
conjunctions add universal ones, each of which must be followed.
In Section~\ref{sec:susana},
we assume formulas in $B^+(Q)$ to be in disjunctive normal form (DNF) and represent them as sets of sets of literals;
hence, $\{\emptyset\}$ and $\emptyset$ stand for $\top$ and $\bot$, respectively.

 \section{\LDLf\ to \AFW}\label{sec:ldlf2afw}

This section describes a translation of dynamic formulas in \LDLf\ to \AFW\ due to \cite{giavar15a}.
More precisely, 
it associates a dynamic formula $\varphi$ in negation normal form with an \AFW\ $\automata_\varphi$,
whose number of states is linear in the size of $\varphi$ and
whose language $\mathcal{L}(\automata_\varphi)$ coincides with the set of all traces satisfying $\varphi$.
A dynamic formula $\varphi$ can be put in negation normal form $\nnf{\varphi}$ by exploiting equivalences and pushing negation inside, until it is only in front of propositional formulas.

The states of $\automata_\varphi$ correspond to the members of the closure \cl{\varphi} of $\varphi$
defined as the smallest set of dynamic formulas such that~\cite{fislad79a}
\begin{enumerate}
\item $\varphi \in \cl{\varphi}$
\item if $\psi \in \cl{\varphi}$ and  $\psi$ is not of the form $\neg \psi'$ then $\neg \psi \in \cl{\varphi}$
\item if $\deventually{\rho}\psi \in \cl{\varphi}$ then $\psi\in \cl{\varphi}$
\item if $\deventually{\psi?}\psi \in \cl{\varphi}$ then $\psi \in \cl{\varphi}$
\item if $\deventually{\rho_1;\rho_2}\psi \in \cl{\varphi}$ then $\deventually{\rho_1}\deventually{\rho_2}\psi \in \cl{\varphi}$
\item if $\deventually{\rho_1+\rho_2}\psi \in \cl{\varphi}$ then $\deventually{\rho_1}\psi \in \cl{\varphi}$ and $\deventually{\rho_2}\psi \in \cl{\varphi}$
\item if $\deventually{\rho^*}\psi \in \cl{\varphi}$ then $\deventually{\rho}\deventually{\rho*}\psi \in \cl{\varphi}$
\end{enumerate}

The alphabet of an \AFW\ $\automata_\varphi$ for a formula $\varphi$ over \PV\ is $\Sigma=2^{\PV\cup\{\last\}}$.
It relies on a special proposition $\last$~\cite{giavar15a}, which is only satisfied by the last state of the trace.
A finite word over $\Sigma$ corresponds to a finite trace over $\PV\cup\{\last\}$.
\begin{definition}[\LDLf\ to \AFW \cite{giavar15a}]
  
Given a dynamic formula $\varphi$ in negation normal form, the corresponding $\AFW$ is defined as
\[
  \automata_\varphi =
  (
  2^{\PV \cup \{\last\}},
  \{\s{\nnf{\phi}} \mid \phi \in \cl{\varphi} \},
  \s{\varphi},
  \delta,
  \emptyset
  )
\]
where transition function $\delta$ mapping
a state $\s{\nnf{\phi}}$ for $\phi \in \cl{\varphi}$ and
an interpretation $\interp\subseteq{\PV\cup\{\last\}}$ into
a positive Boolean formula over the states in $\{\s{\nnf{\phi}} \mid \phi \in \cl{\varphi} \}$
is defined as follows:
\begin{multicols}{2}
\begin{enumerate}
  \item[1.] $\delta(\s{\top},\interp)\eqdef\tpbf $
  \item[3.] $\delta(\s{a},\interp)\eqdef
  \begin{cases}
    \tpbf & \text{if } a   \in\interp    \\
    \fpbf & \text{if } a\notin\interp
  \end{cases}$
  
  \item[5.] $\delta(\s{\deventually{\stp}\varphi},\interp)\eqdef
    \begin{cases}
      \s{\varphi} & \text{if } \last\notin\interp \\
      \fpbf       & \text{if } \last   \in\interp
    \end{cases}$
  \item[2.] $\delta(\s{\bot},\interp)\eqdef\fpbf $
  \item[4.] $\delta(\s{\neg a},\interp)\eqdef
  \begin{cases}
    \fpbf & \text{if } a   \in\interp \\
    \tpbf & \text{if } a\notin\interp
  \end{cases}$

  \item[6.] $\delta(\s{\dalways{\stp}\varphi},\interp)\eqdef
    \begin{cases}
      \s{\varphi} & \text{if } \last\notin\interp \\
      \tpbf       & \text{if } \last   \in\interp
    \end{cases}$
\end{enumerate}
\end{multicols}
\begin{enumerate}
  \setcounter{enumi}{6}
  \item$\delta(\s{\deventually{\psi ?}        \varphi},\interp) \eqdef
  \delta(\s{\psi},\interp) \wedge\delta(\s{\varphi},\interp) $

\item$\delta(\s{\deventually{\rho_1+\rho_2} \varphi},\interp) \eqdef
  \delta(\s{\deventually{\rho_1}\varphi},\interp) \vee  \delta(\s{\deventually{\rho_2}\varphi},\interp) $
\item$\delta(\s{\deventually{\rho_1;\rho_2} \varphi},\interp) \eqdef
  \delta(\s{\deventually{\rho_1}\deventually{\rho_2}\varphi},\interp) $

\item$\delta(\s{\deventually{\rho^{\ast}}    \varphi},\interp) \eqdef
\begin{cases}
  \delta(\s{\varphi},\interp)       & \text{if }  \rho\; \text{is a test}\\
  \delta(\s{\varphi},\interp) \vee  \delta(\s{\deventually{\rho}\deventually{\rho^{\ast}}\varphi},\interp)  & \text{otherwise} \\
\end{cases}  $
\item$\delta(\s{\deventually{(\psi ?)^{\ast}}\varphi},\interp) \eqdef
  \delta(\s{\varphi},\interp) $

\item$\delta(\s{\dalways{\psi ?}            \varphi},\interp) \eqdef
  \delta(\s{\nnf{\neg\psi}},\interp)  \vee  \delta(\s{\varphi},\interp) $
\item$\delta(\s{\dalways{\rho_1 + \rho_2}   \varphi},\interp) \eqdef
  \delta(\s{\dalways{\rho_1}\varphi},\interp) \wedge\delta(\s{\dalways{\rho_2}\varphi},\interp) $

\item$\delta(\s{\dalways{\rho_1 ; \rho_2}   \varphi},\interp) \eqdef
  \delta(\s{\dalways{\rho_1}\dalways{\rho_2}\varphi},\interp) $
  \item$\delta(\s{\dalways{\rho^{\ast}}        \varphi},\interp) \eqdef 
  \begin{cases}
  \delta(\s{\varphi},\interp)       & \text{if }  \rho\; \text{is a test}\\
  \delta(\s{\varphi},\interp) \wedge\delta(\s{\dalways{\rho}\dalways{\rho^{\ast}}\varphi},\interp)  & \text{otherwise} \\
\end{cases}  $

  \item$\delta(\s{\dalways{\rho^{\ast}}        \varphi},\interp) \eqdef
  \delta(\s{\varphi},\interp)  \wedge\delta(\s{\dalways{\rho}\dalways{\rho^{\ast}}\varphi},\interp) $

\item$\delta(\s{\dalways{(\psi ?)^{\ast}}    \varphi},\interp) \eqdef
  \delta(\s{\varphi},\interp) $
\end{enumerate}
\end{definition}

Note that the resulting automaton lacks final states.
This is compensated by the dedicated proposition \last.
All transitions reaching a state, namely $\delta(\s{\dalways{\stp}\varphi},\interp)$ and $\delta(\s{\deventually{\stp}\varphi},\interp)$,
are subject to a condition on \last.
So, for the last interpretation $\interp\cup\{\last\}$,
all transitions end up in \tpbf\ or \fpbf.
Hence, for acceptance, it is enough to ensure that branches reach \tpbf.

As an example,
consider the formula, $\varphi$,
\begin{align}\label{eq:main-example}
  \runexample & = \alwaysF b \wedge \Next a,
\end{align}
stating that $b$ always holds and $a$ is true at the next step.
The \AFW\ for $\varphi$ is $\automata_\varphi =(2^{\{a,b,\last\}},Q^+\cup Q^-,\delta,\emptyset)$,
where
\[
  Q^+ = \{\s{\runexample},\s{\deventually{(\dalways{\stp^*} b)? \;}\deventually{ \stp} a},\s{\dalways{\stp^*} b},\s{\dalways{\stp}\dalways{\stp^*} b},\s{\stp},\s{b},\s{\deventually{\stp}a},\s{a}\}
\]
and $Q^-$ contains all states stemming from negated formulas in $Q^+$; all these are unreachable in our case.
The alternating automaton can be found in in Figure~\ref{fig:ldlafw}.
\begin{figure}[h]
  \centering
  \begin{tikzpicture}[>=latex',node distance=30pt,on grid,auto,initial text=]
    \tikzstyle{every state}=[font=\footnotesize]
    \tikzstyle{interp}=[font=\scriptsize]
    \tikzstyle{and}=[circle,fill,scale=0.3]

    \node[state,initial]  (s_0) {$\s{\varphi}$};
\node[]            (and)    [right = 55pt of s_0]  {$\forall$};
    \node[state]          (s_a)    [below right = 35pt of and]  {$\s{a}$};
    \node[]           (true_a) [right = 50pt of s_a] {$\forall$};
    \node[state]          (s_b)    [above right = 35pt of and]    {$\s{\alwaysF b}$};
    \node[]          (true_1) [right = 50pt of s_b] {$\forall$};

    \path[->]
    (s_0) edge   node[interp] {$b \wedge \neg \last$} (and)
    (and) edge[bend left=10] (s_b)
          edge[bend right=10] (s_a)
    (s_b) edge[below] node[interp] {$b \wedge \last$} (true_1)
          edge[loop above] node[interp] {$b \wedge \neg last$} (s_b)
    (s_a) edge[below] node[interp] {$a$} (true_a);
\end{tikzpicture}
   \caption{$\automata_\varphi$ showing only the reachable states. The special node type, labeled as $\forall$, represents universal transitions, when the $\forall$-node has no outgoing edges it represents the empty universal constraint $\top$.}
  \label{fig:ldlafw}
\end{figure}
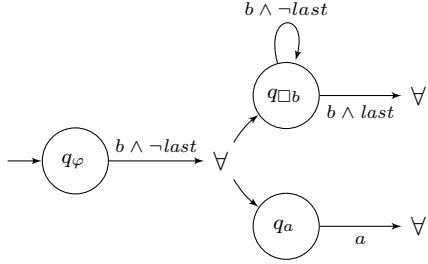

\begin{figure}[h]
  \centering
  \begin{minipage}{.5\textwidth}
    \centering
    \begin{tikzpicture}[minimum width=width=.08cm,font=\footnotesize, shorten >=0.8pt,on grid,auto,node distance=38pt,semithick]
\tikzstyle{interp}=[font=\scriptsize]
\node[](a0) {$\s{\varphi}$};
\node[](a1) [right = 30pt of a0] {$\forall$};
\node[](a2) [above right = 25pt of a1] {$\s{\alwaysF b}$};
\node[](a3) [below right = 25pt of a1] {$\s{a}$};
\node[](a4) [right of = a2] {$\s{\alwaysF b}$};
\node[](a5) [right of = a4] {$\forall \checkmark$};
\node[](a6) [right of = a3] {$\forall \checkmark$};
\path[->]
(a0) edge [dotted] node[above,interp] {$\{b\}$} (a1)
(a1) edge [dotted] node {} (a2)
(a1) edge [dotted] node {} (a3)
(a2) edge [dotted] node[above,interp] {$\{a,b\}$} (a4)
(a4) edge [dotted] node[above,interp] {$\{b,\mathit{last}\}$} (a5)
(a3) edge [dotted] node[above,interp] {$\{a,b\}$} (a6)
;
\end{tikzpicture}
     \caption{Accepted run for \newline $\{b\} \cdot \{a,b\} \cdot \{b, \last\}$.}
    \label{fig:ldlafwrun}
  \end{minipage}\begin{minipage}{.5\textwidth}
    \centering
    \begin{tikzpicture}[minimum width=width=.08cm,font=\footnotesize, shorten >=0.8pt,on grid,auto,node distance=38pt,semithick]
\tikzstyle{interp}=[font=\scriptsize]
\node[](a0) {$\s{\varphi}$};
\node[](a1) [right = 30pt of a0] {$\forall$};
\node[](a2) [above right = 25pt of a1] {$\s{\alwaysF b}$};
\node[](a3) [below right = 25pt of a1] {$\s{a}$};
\node[](a4) [right of = a2] {$\bot$};
\node[](a6) [right of = a3] {$\forall \checkmark$};
\path[->]
(a0) edge [dotted] node[above,interp] {$\{b\}$} (a1)
(a1) edge [dotted] node {} (a2)
(a1) edge [dotted] node {} (a3)
(a2) edge [dotted] node[above,interp] {$\{a\}$} (a4)
(a3) edge [dotted] node[above,interp] {$\{a\}$} (a6)
;
\end{tikzpicture}     \caption{Rejected run for \newline $\{b\} \cdot \{a\} \cdot \{b, \last\}$.}
    \label{fig:ldlafwrun-rejected}
  \end{minipage}
  \end{figure}

\section{Translating \LDLf\ to MSO}\label{sec:ldlf2mso}

It is well-known that while \LTL\ and \LTLf\ can be encoded into first-order logic, the case of \LDL\ and \LDLf\ is rather different.
The encoding of \LDLf\ requires the translation of path expressions of the type $\rho^*$ (the reflexive, transitive closure of a relation), which is not first-order representable.

This is why we need to consider a more expressive formalism and Monadic Second Order ($\mathtt{MSO}$) of Linear Order~\cite{wolfgang97a} ($\mathtt{MSO}(<)$) will be our target logic.
This logic enhances monadic first-order logic of linear order~\cite{kamp68a} with second order quantification. 

\subsection{Monadic Second-order of Linear Order}\label{sec:mso}

Let $\PV$ be an input alphabet
\footnote{By abuse of notation, we use the symbol $\PV$ as for the set of propositional variables in \LDLf, since they will be translated into elements of the alphabet.}
, $\mathcal{V}_1$ be a set of first-order variables denoted by bolded lowercase letters and a set $\mathcal{V}_2$ of second-order variables usually denoted by bolded uppercase letters. 

Well-formed formulas of $\mathtt{MSO}(<)$ are defined according to the following syntax:

\begin{displaymath}
	\varphi:= \fovar{X}(x) \mid \fovar{x} < \fovar{y} \mid \neg \varphi \mid \varphi \vee \psi \mid \exists\; \fovar{x}. \varphi \mid \exists \fovar{X}\; \varphi.
\end{displaymath}

\noindent where $\fovar{x}, \fovar{y} \in \mathcal{V}_1$ and $\fovar{X} \in \mathcal{V}_2$.

The following abbreviations involving logical formulas and orders are valid:

\[
  \arraycolsep=2pt\def\arraystretch{1.3}
  \begin{array}{rclp{10pt}rclp{10pt}rcl}
    \varphi \wedge \psi & \eqdef & \neg \left(\neg \varphi \vee \neg \psi \right)         
	 && \varphi \rightarrow \psi & \eqdef & \neg \varphi \vee \psi
    \\
	\varphi \leftrightarrow \psi & \eqdef & \left(\varphi \rightarrow \psi\right) \wedge \left( \psi \rightarrow \varphi \right)
	  && \forall \fovar{x}\; \varphi & \eqdef & \neg \exists \fovar{x} \;\neg \varphi
	\\
	  \fovar{x} \ge \fovar{y} & \eqdef & \neg \left(\fovar{x} < \fovar{y} \right)
	  && \fovar{x} \le \fovar{y} & \eqdef & \fovar{y} \ge \fovar{x}
	\\
	  \fovar{x} = \fovar{y} & \eqdef & (\fovar{x} \le \fovar{y}) \wedge (\fovar{y} \le \fovar{x})
	  && \fovar{x} \not = \fovar{y} & \eqdef & \neg \left( \fovar{x} = \fovar{y}\right)
	\\ 
	  \fovar{x} > \fovar{y} & \eqdef & \fovar{y} < \fovar{x}
	&&&&

\end{array}
\]

Moreover, the following abbreviations involving first-order and second-order variables will be used along this section. 
\[
  \arraycolsep=2pt\def\arraystretch{1.3}
  \begin{array}{rclp{10pt}rclp{10pt}rcl}
	  \mathtt{succ}(\fovar{x},\fovar{y}) & \eqdef & \fovar{x} < \fovar{y} \wedge \neg \exists \fovar{z} \left(\fovar{x} < \fovar{z} \wedge \fovar{z} < \fovar{y}\right)
\\
	  \mathtt{first}(\fovar{x}) & \eqdef & \neg \exists \fovar{y}\; \fovar{y} < \fovar{x}
	  && \fovar{x} \in \fovar{X} & \eqdef & \fovar{X}(\fovar{x})
	\\
	  \mathtt{last}(\fovar{x}) & \eqdef & \neg \exists \fovar{y}\; \fovar{y} > \fovar{x}
	  && \fovar{X}\not= \fovar{Y} & \eqdef & \neg \fovar{X} = \fovar{Y}
	\\
	  \mathtt{bound}(\fovar{X}, \fovar{w},\fovar{v})& \eqdef & \forall \fovar{r} (\fovar{X}(\fovar{r}) \to (\fovar{w}\leq \fovar{r} \wedge \fovar{r}\leq \fovar{v}))
	  && \fovar{X} = \fovar{Y} & \eqdef & \fovar{X} \subseteq \fovar{Y} \wedge \fovar{Y} \subseteq \fovar{X}
	\\ 
	  \fovar{X} \subseteq \fovar{Y} & \eqdef & \forall \fovar{x} \left(\fovar{x} \in \fovar{X} \rightarrow \fovar{x} \in \fovar{Y}\right)
	&&&&
\end{array}
\]

A $\mathtt{MSO}(<)$ formula is interpreted over a trace $\T$ of length $\lambda$ with respect to two assignments $v_1: \mathcal{V}_1 \mapsto \lbrace 0, \cdots, \lambda -1 \rbrace$ and $v_2: \mathcal{V}_2 \mapsto 2^{\lbrace 0, \cdots, \lambda -1 \rbrace}$. Notice that $v_1$ maps every first-order variable in $\mathcal{V}_1$ into a position in $\T$ while $v_2$ maps each second order variable of $\mathcal{V}_2$ to a set of positions in $\T$.
Given a second-order assignment $v_2$, by $v_2[\fovar{X}:=D]$ we refer to an extension of $v_2$ obtained by assigning to the second-order variable $\fovar{X}$ the set $D \subseteq \lbrace 0, \cdots, \lambda -1 \rbrace$.
For the case of a first-order assignment $v_1$, by $v_1[\fovar{x}:=d]$ we refer to an extension of $v_1$ obtained by assigning to the first-order variable $\fovar{x}$ the value $d \in \lbrace 0, \cdots, \lambda -1 \rbrace$.

\begin{definition}[$\mathtt{MSO}(<)$  satisfaction]
	
A trace $\T$ of length $\lambda$ satisfies a $\mathtt{MSO}(<)$ formula $\varphi$ wrt. assignments $v_1$ and $v_2$, written as $\T, v_1, v_2 \models \varphi$ if the following conditions hold:
\begin{enumerate}
	\item $\T, v_1, v_2 \models \fovar{X}(\fovar{x})$ iff  $v_1(\fovar{x}) \in v_2(\fovar{X})$  \item $\T, v_1, v_2 \models \fovar{x} < \fovar{y}$ iff $v_1(\fovar{x}) < v_1(\fovar{y})$ holds
	\item  $\T, v_1, v_2 \models \neg \varphi$ iff  $\T, v_1, v_2 \not \models \varphi$ 
	\item  $\T, v_1, v_2 \models \varphi \wedge \psi$ iff  $\T, v_1, v_2 \models \varphi$  and $\T, v_1, v_2 \models \psi$
	\item  $\T, v_1, v_2 \models \varphi \vee \psi$ iff  $\T, v_1, v_2 \models \varphi$  or $\T, v_1, v_2 \models \psi$
	\item  $\T, v_1, v_2 \models \varphi \rightarrow \psi$ iff  $\T, v_1, v_2 \not \models \varphi$  or $\T, v_1, v_2 \models \psi$
	\item  $\T, v_1, v_2 \models \exists \fovar{x}\; \varphi$ iff  $\T, v_1[\fovar{x} := d], v_2 \models \varphi$  for some $0 \le d < \lambda$ 
	\item  $\T, v_1, v_2 \models \forall \fovar{x}\; \varphi$ iff  $\T, v_1[\fovar{x}:=d], v_2 \models \varphi$  for all $0 \le d < \lambda$
	\item  $\T, v_1, v_2 \models \exists \fovar{X}\; \varphi$ iff  $\T, v_1, v_2[\fovar{X}:=D] \models \varphi$  for some $D \in 2^{\lbrace0 \cdots \lambda-1\rbrace}$
	\item  $\T, v_1, v_2 \models \forall \fovar{X}\; \varphi$ iff  $\T, v_1, v_2[\fovar{X}:=D] \models \varphi$  for all $D \in 2^{\lbrace 0 \cdots \lambda-1\rbrace}$
\end{enumerate}

\end{definition}

\subsection{Standard Translation}

In this subsection we extend the so called \emph{standard translation} of \LTLf~\cite{giavar13a} to the case of \LDLf.
In order to represent $\rho^*$, first-order logic can be equipped with countable infinite disjunctions as proposed in~\cite{ohnoriga01a}.
Conversely, we use a second order quantified predicate $X$ to capture the points where path expressions are satisfied in a trace.
As in~\cite{ohnoriga01a} our standard translation is defined in terms of two translations $\srtm$ and $\srtp$ for dynamic formulas and paths, respectively.

\begin{definition}[$\LDLf$ Standard Translation]

Given a dynamic formula $\varphi$ and a free variable $\fovar{w}$ representing the time point in which $\varphi$ is evaluated, $\srtm$ is defined as follows: 
\begin{enumerate}
	
	\item $\srtm(\fovar{w},p) \eqdef \; \fovar{P}(\fovar{w}) $
	\item $\srtm(\fovar{w},\top) \eqdef \; \top $
	\item $\srtm(\fovar{w},\bot) \eqdef \; \bot $
	\item $\srtm(\fovar{w},\dalways{\rho}\varphi) \eqdef \; \forall \fovar{v} (\srtp(\fovar{w}\fovar{v},\rho) \to \srtm(\fovar{v},\varphi)) $
	\item $\srtm(\fovar{w},\deventually{\rho}\varphi) \eqdef \; \exists \fovar{v} (\srtp(\fovar{w}\fovar{v},\rho) \wedge \srtm(\fovar{v},\varphi)) $
\end{enumerate}

\noindent The translation for paths $\srtp$ takes as inputs a path expression $\rho$ and two free variables $\fovar{w}$ and $\fovar{v}$ meaning that $\rho$ is satisfied between the time  points $\fovar{w}$ and $\fovar{v}$, defined as follows:
\begin{enumerate}
	\setcounter{enumi}{5}
\item$\srtp(\fovar{w}\fovar{v},\stp) \eqdef  \; \fovar{v}=\fovar{w}+1$
\item$\srtp(\fovar{w}\fovar{v},\varphi ?) \eqdef  \; \srtm(\fovar{w},\varphi) \wedge \fovar{w}=\fovar{v}$
\item$\srtp(\fovar{w}\fovar{v},\rho_1 + \rho_2) \eqdef  \; \srtp(\fovar{w}\fovar{v},\rho_1) \vee \srtp(\fovar{w}\fovar{v},\rho_2)$
\item$\srtp(\fovar{w}\fovar{v},\rho_1;\rho_2) \eqdef  \; \exists \fovar{u} (\srtp(\fovar{w}\fovar{u},\rho_1) \wedge \srtp(\fovar{u}\fovar{v},\rho_2))$
\item$\srtp(\fovar{w}\fovar{v},\rho^\ast) \eqdef  \; \exists \fovar{X} ( \fovar{X}(\fovar{w}) \wedge \fovar{X}(\fovar{v}) \wedge \mathtt{bound}(\fovar{X}, \fovar{w},\fovar{v}) \wedge \mathtt{regular}(\fovar{X}))$
\end{enumerate}

\noindent where $\mathtt{bound}(\fovar{X}, \fovar{w},\fovar{v})$ is defined in Subsection~\ref{sec:mso} and 
	$$\mathtt{regular}(\fovar{X}) \eqdef \forall \fovar{x},\fovar{y}\; \left((\mathtt{succ}(\fovar{x},\fovar{y}) \wedge \fovar{X}(\fovar{x}) \wedge \fovar{X}(\fovar{y})) \to \srtp(\fovar{x}\fovar{y},\rho)\right).$$
\end{definition}

Let $\T$ be a trace of length $\lambda$ and let $v_2$ be an assignment such that each $\fovar{p} \in \PV$, $v_2(\fovar{P}) = \lbrace x \mid  p \in T_x\rbrace$.
With this definition we can prove the model correspondence stated in the following theorem.

\begin{theorem}\label{theorem_ldl2stm}
	Let $\varphi$ be a dynamic formula. Then, for any trace $\T$ of length $\lambda$; $k, d \in [0,\lambda)$ and free variables $\fovar{x}, \fovar{y}$, we have
  \begin{enumerate}
	  \item $\T,k\models \varphi$ iff $\T, v_1[\fovar{x}:=k], v_2\models \srtm(\fovar{x},\varphi)$
	  \item $(k,d) \in \Rel{\rho}{\T}$ iff $\T, v_1[\fovar{x}:=k,\fovar{y}:=d], v_2 \models \srtp(\fovar{x}\fovar{y},\rho)$
  \end{enumerate}
     
\end{theorem}

As an example, the standard translation of the formula $\varphi = \deventually{(\dalways{\stp^*} b)? \; ; \; \stp} a$ with respect to the free variable $\fovar{t}$ is

\begin{alignat*}{4}
	&\srtm(\fovar{t},\varphi)=&& \exists \fovar{v_0} (\exists \fovar{v_1} ( \forall \fovar{v_2} ( &&  \exists \fovar{X} ( \fovar{X}(\fovar{t}) \wedge \fovar{X}(\fovar{v_2}) \wedge \mathtt{bound}(\fovar{X},\fovar{t},\fovar{v_2}) \wedge \\
	&                 &&                       					&& \qquad \forall \fovar{x}, \fovar{y} ((\fovar{X}(\fovar{x})  \wedge \fovar{X}(\fovar{y})  \wedge  \mathtt{succ}(\fovar{x},\fovar{y}))\\
	&                 &&                       && \qquad \qquad \to  ( \fovar{y} = \fovar{x}+1) )) \\
	&                 &&                       && \to \fovar{b}(\fovar{v_2}) )  \quad ( \fovar{v_0} = \fovar{v_1}+1)) \wedge \fovar{a}(\fovar{v_0}))          \\
\end{alignat*}

\subsection{Monadic Second Order Encoding}

In this subsection we provide an alternative translation into ($\mathtt{MSO}(<)$ where the second-order encoding consists of a sequence of existential monadic second-order quantifiers followed by a single universal first-order quantifier. We extend the translation in~\cite{zhpuva19a} from \LTLf\ to the case of \LDLf\, based on the notion of Fisher-Ladner closure~\cite{fislad79a}.

\begin{definition}[$\LDLf$ Monadic Second Order Encoding]
	Given a dynamic formula $\varphi$ and a free variable $\fovar{t}$, $\mso(\fovar{t},\varphi)$ states the truth of $\varphi$ at $\fovar{t}$  as follows:

\begin{align*}
	\mso(\fovar{t}, \varphi) \; \eqdef \; (\exists \fovar{Q_{\theta_0}} \cdots \fovar{\exists Q_{\theta_m}} (\fovar{Q_{\varphi}}(\fovar{t}) \wedge (\forall \fovar{x} (\land_{i=0}^m t(\theta_i,\fovar{x}) ))))
\end{align*}
where, 
$\theta_i \in cl(\varphi)$, 
$\fovar{Q_{\theta_i}}$ is a fresh predicate name and 
$t(\theta_i,\fovar{x})$ asserts the truth of every non-atomic subformula $\theta_i$ in $\cl{\varphi}$ at time point $\fovar{x}$, imitating the semantics of \LDLf, provided in Table~\ref{table:mso:translation}. \end{definition}

\begin{table}[th]
  \begin{center}
    \begin{tabular}{|l|l|}
  \cline{1-2}
$\bm{\mu \in cl(\varphi)}$ & $\bm{t(\mu,\fovar{x})}$\\
  \cline{1-2}
	$\neg \psi$ & $\fovar{Q_{\mu}}(\fovar{x}) \leftrightarrow \neg  \fovar{Q_{\psi}}(\fovar{x})$\\
  \cline{1-2}
	$\deventually{\stp}\psi$ & $\fovar{Q_{\mu}}(\fovar{x}) \leftrightarrow (\exists \fovar{y} (\fovar{y}=\fovar{x}+1 \wedge (\fovar{Q_\psi}(\fovar{y}))) )$\\
  \cline{1-2}
	$\deventually{\mu ?}\psi$ & $ \fovar{Q_{\mu}}(\fovar{x}) \leftrightarrow (\fovar{Q_{\mu}}(\fovar{x}) \wedge \fovar{Q_{\mu}}(\fovar{x}))$\\
  \cline{1-2}
	$\deventually{\rho_1 + \rho_2}\psi$ & $ \fovar{Q_{\mu}}(\fovar{x}) \leftrightarrow
					(\fovar{Q_{\deventually{\rho_1}\psi}}(\fovar{x}) \vee \; \fovar{Q_{\deventually{\rho_2}\psi}}(\fovar{x}))$\\
  \cline{1-2}
	$\deventually{\rho_1 ; \rho_2}\psi$ & $ Q_{\mu}(\fovar{x}) \leftrightarrow
					(\fovar{Q_{\deventually{\rho_1}\deventually{\rho_2}\psi}}(\fovar{x}))$\\
  \cline{1-2}
	$\deventually{\rho^{\ast}}\psi$ & $ \fovar{Q_{\mu}}(\fovar{x}) \leftrightarrow (\fovar{Q_\psi}(\fovar{x}) \vee \fovar{Q_{\deventually{\rho}\deventually{\rho^{\ast}}\psi}}(\fovar{x}))$\\
  \cline{1-2}
	$\dalways{\stp}\psi$ & $\fovar{Q_{\mu}}(\fovar{x}) \leftrightarrow (\forall \fovar{y} (\fovar{y}=\fovar{x}+1 \to (\fovar{Q_\psi}(\fovar{y}))))$\\
  \cline{1-2}
	$\dalways{\mu ?}\psi$ & $ \fovar{Q_{\mu}}(\fovar{x}) \leftrightarrow (\fovar{Q_{\mu}}(\fovar{x}) \to \fovar{Q_{\mu}}(\fovar{x}))$\\
  \cline{1-2}
	$\dalways{\rho_1 + \rho_2}\psi$ & $ \fovar{Q_{\mu}}(\fovar{x}) \leftrightarrow
				    (\fovar{Q_{\dalways{\rho_1}\psi}}(\fovar{x}) \wedge \; \fovar{Q_{\dalways{\rho_2}\psi}}(\fovar{x}))$\\
  \cline{1-2}
	$\dalways{\rho_1 ; \rho_2}\psi$ & $ \fovar{Q_{\mu}}(\fovar{x}) \leftrightarrow
				    (\fovar{Q_{\dalways{\rho_1}\dalways{\rho_2}\psi}}(\fovar{x}))$\\
  \cline{1-2}
	$\dalways{\rho^{\ast}}\psi$ & $ \fovar{Q_{\mu}}(\fovar{x}) \leftrightarrow (\fovar{Q_\psi}(\fovar{x}) \wedge \fovar{Q_{\dalways{\rho}\dalways{\rho^{\ast}}\psi}}(\fovar{x}))$\\
  \cline{1-2}
\end{tabular}
   \end{center}
  \caption{MSO Translation for subformulas in the closure.\label{table:mso:translation}}

\end{table}

Let $\T$ be a trace of length $\lambda$ and let $v_2$ be a second-order assignment such that for each $p \in \PV$, $v_2(\fovar{P}) = \lbrace x \mid  p \in T_x\rbrace$.
With this definition we can prove the model correspondence stated in the following theorem.

\begin{theorem}\label{theorem_ldl2mso}
	Let $\varphi$ be a dynamic formula. Then, for any trace $\T$ of length $\lambda$ and time point $k \in [0,\lambda)$, we have that $\T,k\models \varphi$ iff $\T, v_1[\fovar{t}:=k], v_2 \models \mso(\fovar{t}, \varphi)$. 
\end{theorem}

For instance, given $\varphi = \deventually{(\dalways{\stp^*} b)? \; ; \; \stp} a$, $$\mso(\fovar{t}, \varphi)= \exists\; \fovar{Q_0} \exists  \fovar{Q_1}\; \exists \fovar{Q_2}\;\exists  \fovar{Q_3}\;\exists \fovar{Q_4}\;( \fovar{Q_0}(\fovar{t}) \wedge \forall \fovar{x}\; (\land_{i=0}^4 t(\mu_i,\fovar{x}) ),$$ where each $t(\mu_i,\fovar{x})$ is defined in Table~\ref{table:mso:example}.

\begin{table}[h!]
	\begin{center}
		\begin{tabular}{|c|l|l|}
  \cline{1-3}
  $\bm{Q_\mu}$ & $\bm{\mu \in cl(\varphi)}$ & $\bm{t(\mu,\fovar{x})}$\\
  \cline{1-3}
$\fovar{Q_0}$ & $\deventually{(\dalways{\stp^*} b)? \; ; \; \stp} a$ & $\fovar{Q_0}(\fovar{x}) \leftrightarrow \fovar{Q_1}(\fovar{x})$\\
  \cline{1-3}
$\fovar{Q_1}$ & $\deventually{(\dalways{\stp^*} b)?}\deventually{\stp} a$ & $\fovar{Q_1}(\fovar{x}) \leftrightarrow \fovar{Q_4}(\fovar{x}) \wedge \fovar{Q_2}(\fovar{x})$ \\
  \cline{1-3}
	$\fovar{Q_2}$ & $\dalways{\stp^*} b$ & $\fovar{Q_2}(\fovar{x}) \leftrightarrow \fovar{Q_b}(\fovar{x}) \wedge \fovar{Q_3}(\fovar{x}) $\\
  \cline{1-3}
	$\fovar{Q_3}$ & $\dalways{\stp}\dalways{\stp^*} b$ & $\fovar{Q_3}(\fovar{x}) \leftrightarrow \forall \fovar{v} \; \fovar{v}=\fovar{x}+1 \to \fovar{Q_2}(\fovar{v})$\\
  \cline{1-3}
	$\fovar{Q_4}$ & $\deventually{\stp} a$ & $\fovar{Q_4}(\fovar{x}) \leftrightarrow \exists \fovar{v} \; \fovar{v}=\fovar{x}+1 \wedge \fovar{Q_a}(\fovar{v})$ \\
  \cline{1-3}
\end{tabular}
 		\caption{MSO Translation for non atomic subformulas of $\varphi$.\label{table:mso:example}}
	\end{center}
\end{table}

\section{Using automata for implementing dynamic constraints}
\begin{figure}[!h]
    \includegraphics[width=\textwidth]{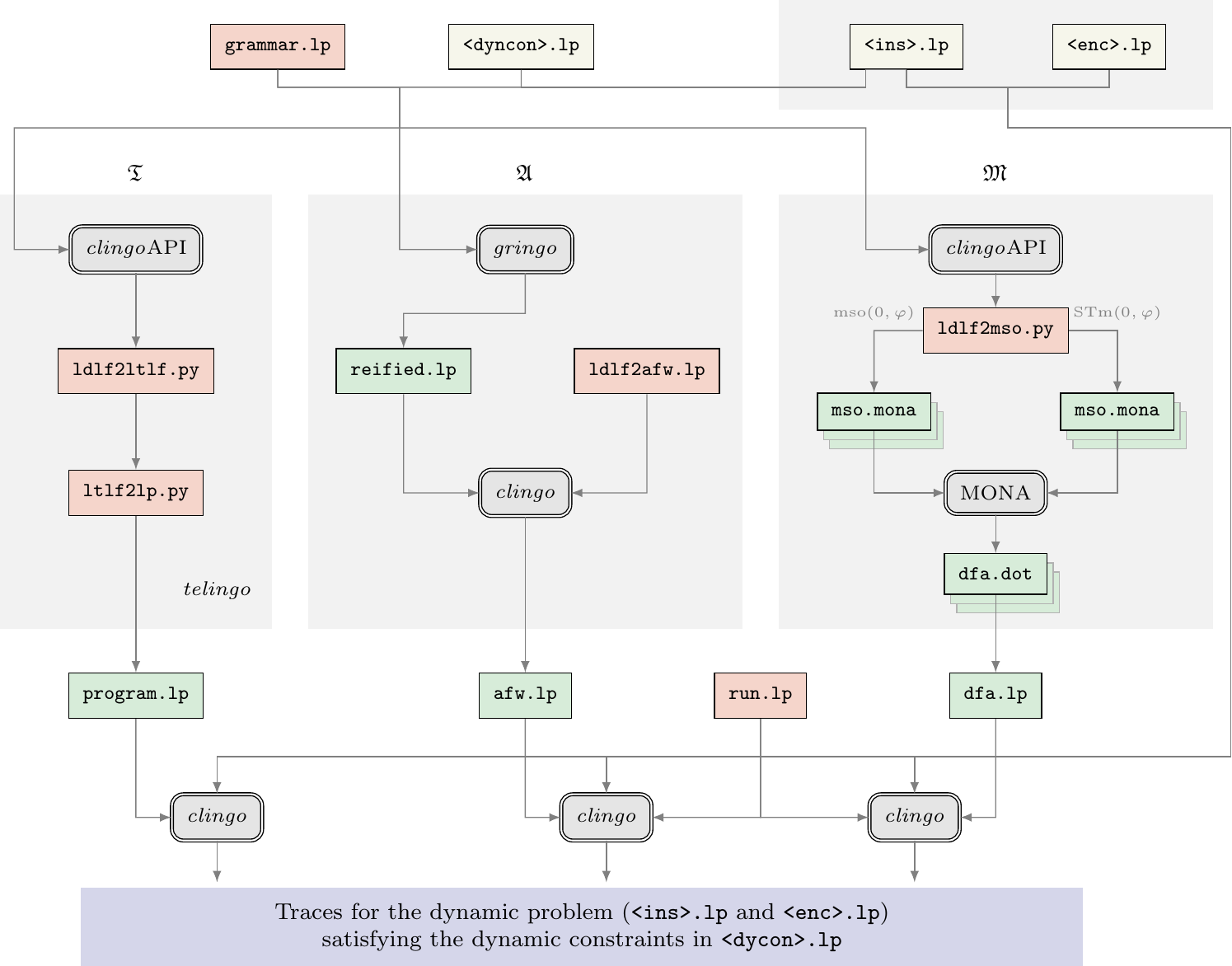}
\caption{Workflows of our framework.
      Elements in
      yellow correspond to user input,
      green ones are automatically generated, and
      red ones are provided by the system to solve the problem.}
    \label{fig:workflow}
\end{figure}
Our goal is to investigate alternative ways of implementing constraints imposed by dynamic formulas.
To this end,
we pursue three principled approaches:
\begin{enumerate}
\item[(\WFT)] Tseitin-style translation into regular logic programs,
\item[(\WFA)] ASP-based translation into alternating automata,
\item[(\WFM)] MONA-based translation into deterministic automata, using \WFMm\ and \WFMs\ for the Monadic Second Order Encoding and the Standard Translation, respectively.
\end{enumerate}
These alternatives are presented in our systems' workflow \footnote{The source code can be found in \url{https://github.com/potassco/atlingo} v1.0.} from Figure~\ref{fig:workflow}.
The common idea is to compute all fixed-length traces, or plans, of a dynamic problem expressed in
plain ASP (in files \texttt{<ins>.lp} and \texttt{<enc>.lp}) that satisfy the
dynamic constraints in \texttt{<dyncon>.lp}.
All such constraints are of form \texttt{:- not $\mathtt{\varphi}$.} which is the logic programming representation of the formula $\neg \neg \varphi$.
Note that these constraints may give rise to even more instances after grounding.
The choice of using plain ASP rather than temporal logic programs, as used in \telingo~\cite{cakascsc18a,cakamosc19a},
is motivated by simplicity and the possibility of using existing ASP benchmarks.

For expressing dynamic formulas all three approaches rely on \clingo's theory reasoning framework that allows for
customizing its input language with theory-specific language constructs that are defined by a theory grammar~\cite{kascwa17a}.
The part \telingo\ uses for dynamic formulas is given in Listing \ref{lst:theory:grammar}.
\begin{lstlisting}[label={lst:theory:grammar},caption={Theory specification for dynamic formulas (\texttt{grammar.lp})},captionpos=b,language=clingo,basicstyle=\ttfamily\footnotesize]
#theory del {
 formula_body {
    & : 7,unary; ~ : 5,unary;%%#(\label{gra:ops:boolean}#)
    ? : 4,unary; * : 3,unary; + : 2,binary,left; ;; : 1,binary,left;%%#(\label{gra:ops:path}#)
  .>? : 0,binary,right;     .>* : 0,binary,right%%#(\label{gra:ops:dynamic}#)
 };
 &del/0 : formula_body, body%%#(\label{gra:formula:body}#)
}.
\end{lstlisting}
The grammar contains a single theory term definition for \lstinline{formula_body},
which consists of terms formed from the theory operators in Line~\ref{gra:ops:boolean} to~\ref{gra:ops:dynamic}
along with basic \gringo\ terms.
More specifically,
\lstinline{&} serves as a prefix for logical constants, eg.\ \lstinline{&true} and \lstinline{&t} stand for $\top$ and \stp,
while \lstinline{~} stands for negation.
The path operators $?$, ${}^*$, $+$, $;$ are represented by \lstinline{?}, \lstinline{*}, \lstinline{+}, \lstinline{;;},
where \lstinline{?} and \lstinline{*} are used as prefixes,
and
the binary dynamic operators \deventually{\cdot} and \dalways{\cdot} by \lstinline{.>?} and \lstinline{.>*}, respectively
(extending \telingo's syntax \lstinline{>?} and \lstinline{>*} for unary temporal operators \eventuallyF\ and \alwaysF).
Such theory terms can be used within the set associated with the (zero-ary) theory predicate \lstinline{&del/0}
defined in Line~\ref{gra:formula:body} (cf.~\cite{kascwa17a}).
Since we impose our dynamic constraints through integrity constraints,
we restrict the occurrence of corresponding atoms to rule bodies, as indicated by the keyword \lstinline{body}.
The representation of our running example \runexample\ as an integrity constraint is given in Listing~\ref{lst:theory:del-constraint}.
\begin{lstlisting}[label={lst:theory:del-constraint},caption={Representation of `$\leftarrow\neg\runexample$' from~\eqref{eq:main-example} (\texttt{delex.lp})},captionpos=b,language=clingo,numbers=none]
:- not &del{ ? (* &t .>* b) ;; &t .>? a }.
\end{lstlisting}Once such a dynamic formula is parsed by \gringo, it is processed in a different way in each workflow.
At the end, however, each workflow produces a logic program that is combined with the original dynamic problem in
\texttt{<ins>.lp} and \texttt{<enc>.lp} and handed over to \clingo\ to
compute all traces of length \lstinline{lambda} satisfying the dynamic formula(s) in \texttt{<dyncon>.lp}.
We also explored a translation from the alternating automata generated in \WFA\ into an NFA using both ASP and python.
This workflow, however, did not show any interesting results, hence, due to space limitations it is omitted.
 
\subsection{Tseitin-style translation into logic programs}
\label{sec:tseitin}

The leftmost part of the workflow in Figure~\ref{fig:workflow} relies on \telingo's infrastructure~\cite{cakascsc18a,cakamosc19a}:
Once grounded,
a dynamic formula is first translated into a temporal formula (\texttt{ldlf2ltlf.py}),
which is then translated into a regular logic program (\texttt{ltlf2lp.py}).\footnote{Filenames are of indicative nature only.}
These translations heavily rely on the introduction of auxiliary variables for subformulas, a technique due to Tseitin~\cite{tseitin68a}.
In this way, all integrity constraints in \texttt{<dyncon>.lp} get translated into the ground program \texttt{program.lp}.
In the worst case, this program consists of \lstinline{lambda} copies of the translated constraint.
This approach is detailed in~\cite{cakamosc19a,cadilasc20a}.

\subsection{ASP-based translation into alternating automata}
\label{sec:susana}

The approach illustrated in the middle of Figure~\ref{fig:workflow} follows the construction in Section~\ref{sec:ldlf2afw}.
More precisely, it builds the \AFW\ $\automata_\varphi$ for each ground constraint $\neg\neg\varphi$
by taking advantage of Proposition~\ref{pro:dht:ldl}.
Notably, the approach is fully based on ASP and its meta-programming capabilities:
It starts by reifying each $\neg\neg\varphi$ into a set of facts, yielding the single file \texttt{reified.lp}.
These facts are then turned into one or more \AFW\ $\automata_\varphi$ through logic program \texttt{ldlf2afw.lp}.
In fact,
each $\automata_\varphi$ is once more represented as a set of facts, gathered in file \texttt{afw.lp} in Figure~\ref{fig:workflow}.
Finally, the encoding in \texttt{run.lp} makes sure that the trace produced by the encoding of the original dynamic problem
is an accepted run of $\automata_\varphi$.

In what follows, we outline these three steps using our running example.

The dynamic constraint in Listing~\ref{lst:theory:del-constraint} is transformed into a set of facts via \gringo's
reification option \texttt{--output=reify}.
The facts provide a serialization of the constraint's abstract syntax tree following the \aspif\ format~\cite{kascwa17a}.
Among the 42 facts obtained from Listing~\ref{lst:theory:del-constraint},
we give the ones representing subformula $\dalways{\stp^*} b$, or `\lstinline{* &t .>* b}', in Listing~\ref{lst:theory:reified}.
\begin{lstlisting}[float=ht,label={lst:theory:reified},caption={Facts~\lstinline{11}-\lstinline{20} obtained by a call akin to \texttt{gringo --output=reify grammar.lp delex.lp > reified.lp}},firstnumber=11,captionpos=b,language=clingo]
theory_string(5,"*").%%#(\label{rex:reified:fact:one}#)
theory_tuple(1).%%#(\label{rex:reified:fact:two}#)
theory_tuple(1,0,8).%%#(\label{rex:reified:fact:tri}#)
theory_function(9,5,1).%%#(\label{rex:reified:fact:for}#)
theory_string(10,"b").%%#(\label{rex:reified:fact:fiv}#)
theory_string(4,".>*").%%#(\label{rex:reified:fact:six}#)
theory_tuple(2).%%#(\label{rex:reified:fact:svn}#)
theory_tuple(2,0,9).%%#(\label{rex:reified:fact:eit}#)
theory_tuple(2,1,10).%%#(\label{rex:reified:fact:nin}#)
theory_function(11,4,2).%%#(\label{rex:reified:fact:ten}#)
\end{lstlisting}
\Gringo's reification format uses integers to identify substructures and to tie them together.
For instance, the whole expression `\lstinline{* &t .>* b}' is identified by \lstinline{11} in Line~\ref{rex:reified:fact:ten}.
Its operator `\lstinline{.>*}' is identified by \lstinline{4} and both are mapped to each other in Line~\ref{rex:reified:fact:six}.
The two arguments `\lstinline{* &t}' and `\lstinline{b}' are indirectly represented by tuple~\lstinline{2} in
Line~\ref{rex:reified:fact:svn}-\ref{rex:reified:fact:nin} and identified by \lstinline{9} and \lstinline{10}, respectively.
While `\lstinline{b}' is directly associated with \lstinline{10} in Line~\ref{rex:reified:fact:fiv},
`\lstinline{* &t}' is further decomposed in Line~\ref{rex:reified:fact:for} into operator `\lstinline{*}'
(cf.~Line~\ref{rex:reified:fact:one}) and its argument `\lstinline{&t}'.
The latter is captured by tuple~\lstinline{1} but not further listed for brevity.

The reified representation of the dynamic constraint in Listing~\ref{lst:theory:del-constraint} is now used to build
the \AFW\ in Figure~\ref{fig:ldlafw} in terms of the facts in Listing~\ref{lst:automaton}.
\begin{lstlisting}[float=ht,label={lst:automaton},caption={Generated facts representing the \AFW\ in Figure~\ref{fig:ldlafw} (\texttt{afw.lp})},captionpos=b,language=clingo]
prop(10,"b").%%#(\label{aex:prop:one}#)
prop(14,"a").
prop(16,"last").%%#(\label{aex:prop:tri}#)
state(0,dia(seq(test(box(str(stp),p(10))),stp),p(14))).%%#(\label{aex:state:one}#)
state(1,p(14)).%%#(\label{aex:state:tri}#)
state(2,box(str(stp),p(10))).%%#(\label{aex:state:two}#)
initial_state(0).
delta(0,0). delta(0,0,out,16). delta(0,0,in,10).%%#(\label{aex:transition:zro}#)
            delta(0,0,1). delta(0,0,2).%%#(\label{aex:transition:zro:q}#)
delta(1,0). delta(1,0,in,14).%%#(\label{aex:transition:one}#)
delta(2,0). delta(2,0,out,16). delta(2,0,in,10).%%#(\label{aex:transition:two:zro}#)
            delta(2,0,2).%%#(\label{aex:transition:two:zro:q}#)
delta(2,1). delta(2,1,in,16). delta(2,1,in,10).%%#(\label{aex:transition:two:one}#)
\end{lstlisting}
As shown in Figure~\ref{fig:workflow}, the facts in \texttt{afw.lp} are obtained by applying \clingo\ to \texttt{ldlf2afw.lp} and
\texttt{reified.lp}, the facts generated in the first step.

An automaton $\automata_\varphi$ is represented by the following predicates:
\begin{itemize}
\item \lstinline{prop/2}, providing a symbol table mapping integer identifiers to atoms,
\item \lstinline{state/2}, providing states along with their associated dynamic formula;
  the initial state is distinguished by \lstinline{initial_state/1}, and
\item \lstinline{delta/2,3,4}, providing the automaton's transitions.
\end{itemize}
The symbol table in Line~\ref{aex:prop:one} to~\ref{aex:prop:tri} in Listing~\ref{lst:automaton} is directly derived from the reified format.
In addition, the special proposition \lstinline{last} is associated with the first available identifier.
The interpretations over \lstinline{a}, \lstinline{b}, \lstinline{last} constitute the alphabet of the automaton at hand.

More efforts are needed for calculating the states of the automaton.
Once all relevant symbols and operators are extracted from the reified format,
they are used
to build the closure \cl{\varphi} of $\varphi$ in the input and
to transform its elements into negation normal form.
In the final representation of the automaton,
we only keep reachable states and assign them a numerical identifier.
The states in Line~\ref{aex:state:one} to~\ref{aex:state:tri} correspond to the ones labeled
\s{\varphi}, \s{a} and \s{\alwaysF b} in Figure~\ref{fig:ldlafw}.

The transition function is represented by binary, ternary, and quaternary versions of predicate \lstinline{delta}.
The representation is centered upon the conjunctions in the set representation of the DNF of
$\delta(q,\interp)$ (cf.\ Section~\ref{sec:automata}).
Each conjunction \lstinline{C} represents a transition from state \lstinline{Q} and is captuted by \lstinline{delta(Q,C)}.
An atom of form \lstinline{delta(Q,C,Q')} indicates that state \lstinline{Q'} belongs to conjunction \lstinline{C}
and \lstinline{delta(Q,C,T,A)} expresses the condition that either $\mathtt{A}\in\interp$ or $\mathtt{A}\not\in\interp$
depending on whether \lstinline{T} equals \lstinline{in} or \lstinline{out}, respectively.
The binary version of \lstinline{delta} is needed since there may be no instances of the ternary and quaternary ones.

The facts in Line~\ref{aex:transition:zro} to~\ref{aex:transition:zro:q} in Listing~\ref{lst:automaton} capture
the only transition from the initial state in Figure~\ref{fig:ldlafw},
viz.\ $\delta(\s{\varphi},\interp)=\{\{\s{\alwaysF{b}},\s{a}\}\}$.
Both the initial state and the transition are identified by \lstinline{0} in Line~\ref{aex:transition:zro}.
Line~\ref{aex:transition:zro} also gives the conditions $\last\not\in\interp$ and $b\in\interp$ 
needed to reach the successor states given in Line~\ref{aex:transition:zro:q}.
Line~\ref{aex:transition:one} accounts for $\delta(\s{a},\interp)=\{\emptyset\}$,
reaching $\top$ (ie., an empty set of successor states) from \s{a} provided $a\in\interp$.
We encounter two possible transitions from state \lstinline{2}, or \s{\dalways{\stp^*}b}.
Transition \lstinline{0} in Line~\ref{aex:transition:two:zro} to~\ref{aex:transition:two:zro:q} represents the loop
\(
\delta(\s{\dalways{\stp^*}b},\interp)=\{\{\s{\dalways{\stp^*}b}\}\}
\)
for
$\last\not\in\interp$ and $b\in\interp$,
while
transition \lstinline{1} in Line~\ref{aex:transition:two:one} captures
\(
\delta(\s{\dalways{\stp^*}b},\interp)=\{\emptyset\}
\)
that allows us to reach $\top$ whenever $\{\last,b\}\subseteq\interp$.

Finally, the encoding in Listing~\ref{lst:runs} checks whether a trace is an accepted run of a given automaton.
\begin{lstlisting}[float=ht,label={lst:runs},caption={Encoding defining the accepted runs of an automaton \texttt{(run.lp)}.},captionpos=b,language=clingo,basicstyle=\small\ttfamily]
node(Q,0) :- initial_state(Q).%%#(\label{lst:runs:one}#)

{ select(C,Q,T): delta(Q,C) } = 1 :- node(Q,T), T<=lambda-1.%%#(\label{lst:runs:two}#)

node(Q',T+1) :- select(C,Q,T), delta(Q,C,Q').%%#(\label{lst:runs:tri}#)

:- select(C,Q,T), delta(Q,C,in,A), not trace(A,T).%%#(\label{lst:runs:for}#)
:- select(C,Q,T), delta(Q,C,out,A), trace(A,T).%%#(\label{lst:runs:fiv}#)
\end{lstlisting}
We describe traces using atoms of form \lstinline{trace(A,T)}, stating that the atom identified by \lstinline{A} is true in the trace at time step \lstinline{T}.
Although such traces are usually provided by the encoding of the dynamic problem at hand, the accepted runs of
an automaton can also be enumerated by adding a corresponding choice rule.
In addition, the special purpose atom \lstinline{last} is made true in the final state of the trace.

For verifying whether a trace of length \lstinline{lambda} is accepted, we build the tree corresponding to a run of the \AFW\ on the trace at hand.
This tree is represented by atoms of form \lstinline{node(S,T)}, indicating that state \lstinline{S} exists at depth/time \lstinline{T}\footnote{Note that we do not need to represent the edges between nodes as their depth is indicative enough for the acceptance. In the literature, runs of \AFW\ are often represented using directed acyclic graphs instead of trees.}.
The initial state is anchored as the root in Line~\ref{lst:runs:one}.
In turn, nodes get expanded by depth by selecting possible transitions in Line~\ref{lst:runs:two}.
The nodes are then put in place by following the transition of the selected conjunction in Line~\ref{lst:runs:tri}.
Lines~\ref{lst:runs:for} and~\ref{lst:runs:fiv} verify the conditions for the selected transition.

\subsection{MONA-based translation into deterministic automata}
\label{sec:mona}

The rightmost part of the workflow in Figure~\ref{fig:workflow} relies on our translations of dynamic formulas into
MSOs in Section~\ref{sec:ldlf2mso}.
We use the off-the-shelf tool MONA\footnote{\url{https://www.brics.dk/mona}}~\cite{hejejoklparasa95a} to
translate the resulting MSO formulas into DFAs.
More precisely,
we use \clingo's API to transform each dynamic constraint $\neg\neg\varphi$ in \texttt{<dyncon>.lp} either into
MSO formula $\mathit{mso}(0,\varphi)$ or $\mathit{stm}(0,\varphi)$.
This results in a file \texttt{mso.mona} in MONA's syntax,
which is then turned by MONA into a corresponding DFA in \texttt{dot} format.
All these automata are then translated into facts and
gathered in \texttt{dfa.lp} (Listing~\ref{lst:dfa_automaton}) in the same format as used for \AFW{s}.
The encoding in Listing~\ref{lst:runs} can be used to find accepted runs of DFAs by adding the following integrity
constraint ensuring that runs end in a final state.
\begin{lstlisting}[language=clingo,numbers=none]
:- node(Q,lambda), not final_state(Q).
\end{lstlisting}

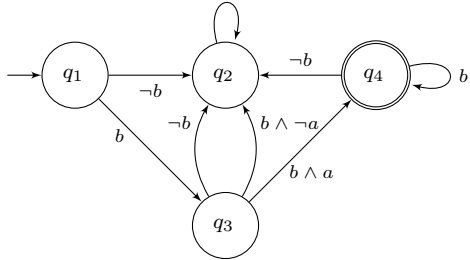
\begin{figure}[h]
    \centering
    \begin{tikzpicture}[>=latex',node distance=2cm,on grid,auto,initial text=]
	\tikzstyle{every state}=[font=\footnotesize]
    \tikzstyle{interp}=[font=\scriptsize]

	\node[state,initial] (a)  {$\s{1}$};
	\node[state] (b) [right  of=a] {$\s{2}$};
	\node[state] (c) [below   of=b] {$\s{3}$};
	\node[state,accepting] (d) [right of=b] {$\s{4}$};
	\path[->]
	(a) edge  node[interp, below] {$\neg b$} (b)
	(a) edge  node[interp, below, pos=.2] {$b$} (c)
	(c) edge[bend left]  node[interp, left, pos=.8] {$\neg b$} (b)
	(c) edge[bend right]  node[interp, right, pos=.8] {$b \wedge \neg a$} (b)
	(c) edge[]  node[interp, right, pos=0.3] {$b \wedge a$} (d)
	(d) edge[]  node[interp, above] {$\neg b$} (b)
	(b) edge[loop above]  node[interp, right] {} (b)
	(d) edge[loop right]  node[interp, right] {$b$} (d);
\end{tikzpicture}

     \caption{DFA automata computed by MONA for $\varphi$ (\ref{eq:main-example}).}
    \label{fig:dfa}
  \end{figure}

\begin{lstlisting}[label={lst:dfa_automaton},caption={Facts representing the DFA in Figure~\ref{fig:dfa} (\texttt{dfa.lp})},captionpos=b,language=clingo]
prop(0,"a").
prop(1,"b").
prop(2,"last").
state(1,"q1").
state(2,"q2").
state(3,"q3").
state(4,"q4").
initial_state(1).
delta(1,0). delta(1,0,2). delta(1,0,out,1).
delta(1,1). delta(1,1,3). delta(1,1,in,1).
delta(2,0). delta(2,0,2).
delta(3,0). delta(3,0,2). delta(3,0,out,0).
delta(3,1). delta(3,1,2). 
            delta(3,1,in,0). delta(3,1,out,1).
delta(3,2). delta(3,2,4). 
            delta(3,2,in,0). delta(3,2,in,1).
delta(4,0). delta(4,0,2). delta(4,0,out,1).
delta(4,1). delta(4,1,4). delta(4,1,in,1).
final_state(4).
\end{lstlisting}

\section{Evaluation}\label{sec:evaluation}

\begin{figure}[h]
  \centering
  \begin{minipage}{.43\textwidth}
    \centering
    \includegraphics[scale=0.15]{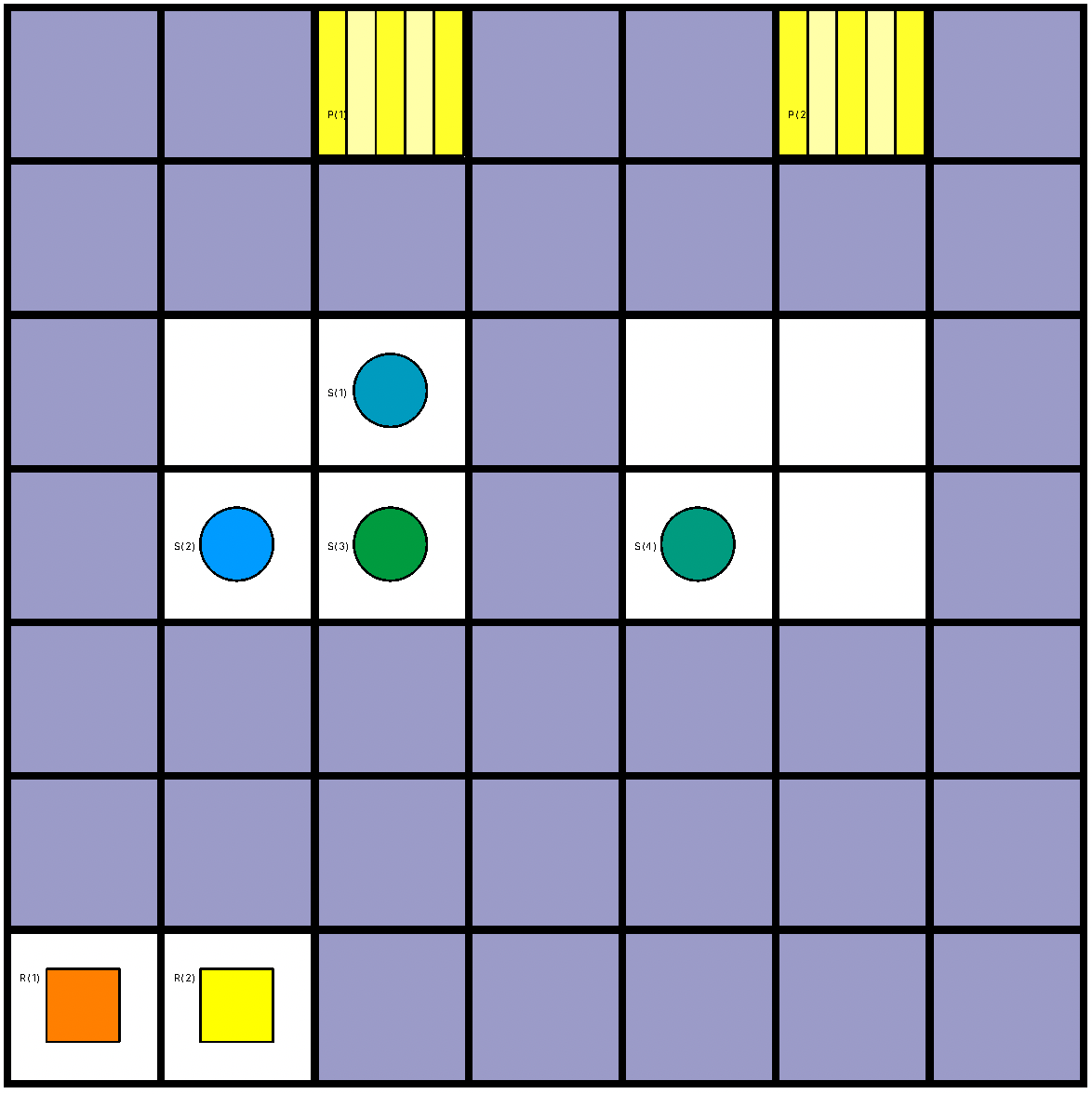}
    \caption{Asprilo visualization for two robots instance.}
    \label{fig:asprilo-2}
  \end{minipage}
  \hspace{0.1\textwidth}
  \begin{minipage}{.43\textwidth}
  \centering
    \includegraphics[scale=0.15]{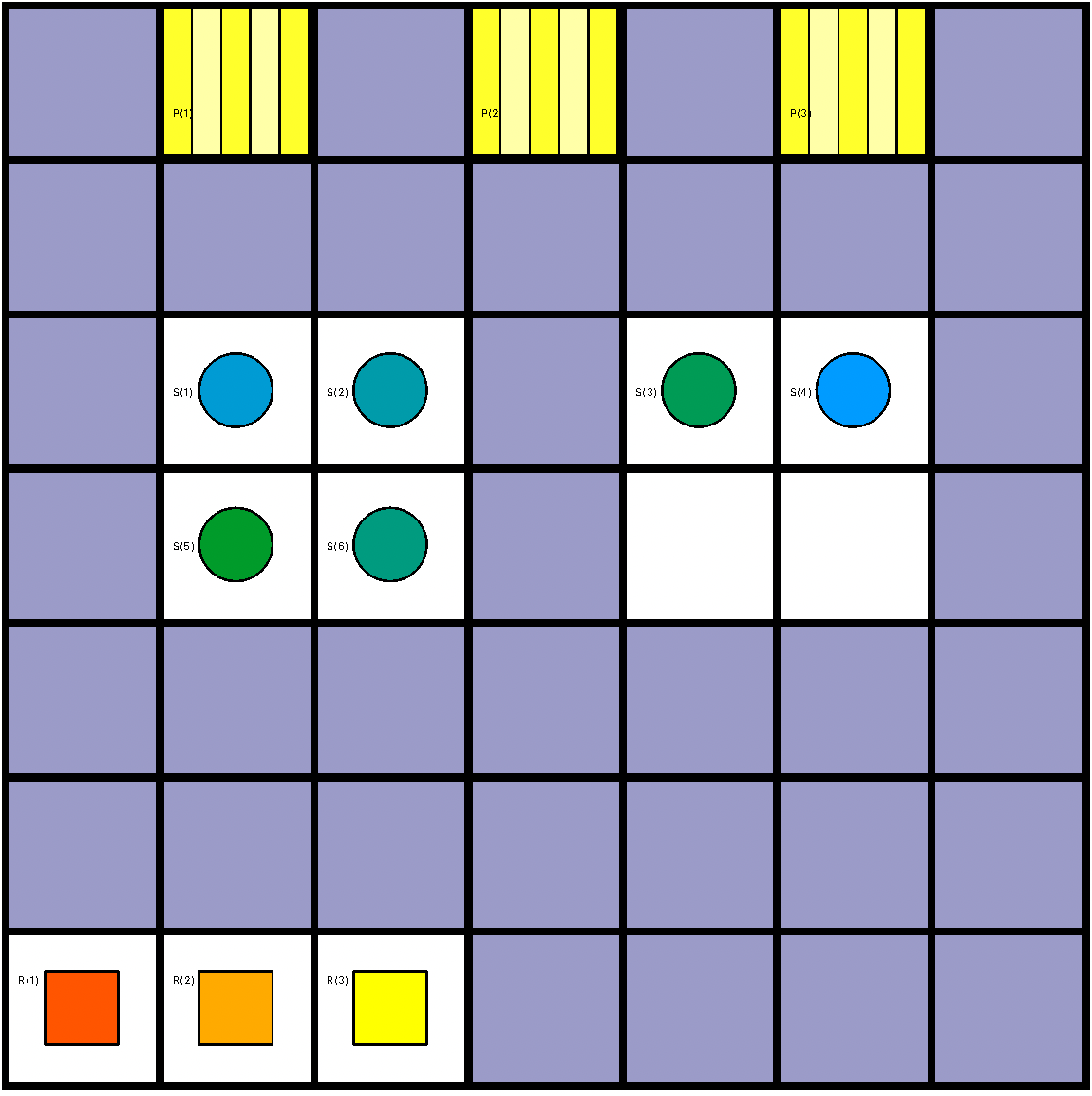}
    \caption{Asprilo visualization for three robots instance.}
    \label{fig:asprilo-3}
  \end{minipage}
\end{figure}

For our experimental studies, we use benchmarks from the domain of robotic intra-logistics
stemming from the \asprilo\ framework~\cite{geobotscsangso18a}.
As illustrated in Figure~\ref{fig:asprilo-2} and~\ref{fig:asprilo-3},
we consider grids of size 7$\times$7 with $n\in\{2,3\}$ robots and $n*2$ orders of single products,
each located on a unique shelf.
At each timestep, a robot can: (i) $\mathit{move}$ in a direction(ii) $\mathit{pickup}$ a shelf (iii) $\mathit{putdown}$ a shelf or (iv) $\mathit{wait}$. 
Moreover, a robot will $\mathit{deliver}$ an order if it waits at a picking station while carrying a shelf.
The goal is to take each shelf to a picking station;
in an optimal plan (wrt. trace length) each robot processes two orders. 

We consider three different dynamic constraints.
The first one restricts plans such that
if a robot picks up a shelf,
then it must move or wait several times
until the shelf is delivered.
This is expressed by the dynamic formula $\varphi_1$ and represented in Listing \ref{lst:evaluation:asprilo-d1}\footnote{We start repetitions with $\stp$ as \texttt{\&t}, to cope with movements in \asprilo\ starting at time point 1.}, were $\mathit{pickup}_s$ and $\mathit{deliver}_s$ refer to a specific shelf.
\begin{align*}
  \varphi_1=\dalways{\stp^\ast} \dalways{\mathit{pickup}_s?} \deventually{(\stp ; (\mathit{move}? + \mathit{wait}?))^\ast ; \mathit{deliver}_s?} \top
\end{align*}
\begin{lstlisting}[float=ht, label={lst:evaluation:asprilo-d1}, language=clingo, caption={Dynamic constraint for formula $\varphi_1$.},captionpos=b,basicstyle=\ttfamily\footnotesize]
:- not &del{
      (* &t) .>* 
      ?pickup(robot(R),shelf(S)) .>* 
      *(&t ;; ?move(robot(R)) + ?waits(robot(R))) ;; 
        ?deliver(robot(R),shelf(S)) .>? 
      &true},
   robot(R), shelf(S). 
\end{lstlisting}
The second one, $\varphi_2$, represents a procedure where
robots must repeat a sequence in which they move towards a shelf, pickup, move towards a picking station, deliver, move
to the dropping place and putdown, and finish with waiting until the end of the trace;
it is represented in Listing~\ref{lst:evaluation:asprilo-d2}.
\begin{align*}
\varphi_2=\deventually{(\mathit{move}^\ast ;\mathit{pickup}^\ast;\mathit{move}^\ast;\mathit{deliver};\mathit{move}^\ast;\mathit{putdown})^\ast ; \mathit{wait}^\ast} \finally
\end{align*}
\begin{lstlisting}[float=ht, label={lst:evaluation:asprilo-d2}, language=clingo, caption={Dynamic constraint for formula $\varphi_2$.},captionpos=b,basicstyle=\ttfamily\footnotesize]
:- not &del{
           *( *(&t ;; ?move(robot(R))) ;; 
              &t ;; ?pickup(robot(R)) ;;  
              *(&t ;; ?move(robot(R))) ;;  
              &t ;; ?deliver(robot(R));; ?waits(robot(R)) ;; 
              *(&t ;; ?move(robot(R)));; 
               &t ;; ?putdown(robot(R))) 
		   ;; *(&t ;; ?waits(robot(R))) 
        .>? &t.>* &false }, robot(R).
\end{lstlisting}
For our last constraint we use the dynamic formula $\varphi_3$ given in Listing~\ref{lst:evaluation:asprilo-d3}.
This corresponds to a procedure similar to $\varphi_2$ but which relies on a predefined pattern, restricting the direction of movements with $\mathit{move}_\mathit{r}$, $\mathit{move}_\mathit{l}$, $\mathit{move}_\mathit{u}$ and $\mathit{move}_\mathit{d}$ to refer to moving  right, left, up and down, respectively. We use the path $\rho=(\mathit{move}_\mathit{r}^{\;\ast} + \mathit{move}_\mathit{l}^{\;\ast})$ so that robots only move in one horizontal direction.
Additionally, each iteration starts by waiting so that whenever a robot starts moving,
it fulfills the delivery without intermediate waiting.
\begin{align*}
\varphi_3=\deventually{(\mathit{wait}^\ast ; \rho ; \mathit{move}_\mathit{u}^{\;\ast} ; \mathit{pickup}; \rho ; \mathit{move}_\mathit{u}^{\;\ast};\mathit{deliver};\rho ; \mathit{move}_\mathit{d}^{\;\ast};\mathit{putdown})^\ast ; \mathit{wait}^\ast} \finally
\end{align*}
\begin{lstlisting}[float=ht, label={lst:evaluation:asprilo-d3}, language=clingo, caption={Dynamic constraint for formula $\varphi_3$.},captionpos=b,basicstyle=\ttfamily\footnotesize]
:- not &del{
    *(  *(&t ;; ?waits(robot(R))) ;;
        (  *(&t ;; (?move(robot(R),RIGHT))) +
            *(&t ;; (?move(robot(R),LEFT)))
        ) ;; 
        *(&t ;; ?move(robot(R),UP) ) ;;

        &t ;; ?pickup(robot(R)) ;; 

        (  *(&t ;; (?move(robot(R),RIGHT))) +
            *(&t ;; (?move(robot(R),LEFT)))
        ) ;; 
        *(&t ;; ?move(robot(R),UP) ) ;;

        &t ;; ?deliver(robot(R));; ?waits(robot(R)) ;;

        (  *(&t ;; (?move(robot(R),RIGHT))) +
            *(&t ;; (?move(robot(R),LEFT)))
        ) ;; 
            *(&t ;; ?move(robot(R),DOWN) ) ;;
        
        &t ;; ?putdown(robot(R)))
    ;; *(&t ;; ?waits(robot(R)))
.>? &t.>* &false }, 
   robot(R), up(UP), right(RIGHT), left(LEFT), down(DOWN).
\end{lstlisting}

We use these constraints to contrast their implementations by means of our workflows
\WFA, \WFT, \WFMm\ and \WFMs\ with $\lambda\in\{25,\dots,31\}$,
while using the option of having no constraint, namely \WFNC, as a baseline.
The presented results ran using \clingo~5.4.0 on an Intel Xeon E5-2650v4 under Debian GNU/Linux~9, with a memory of 20~GB and a timeout of 20~min per instance.
All times are presented in milliseconds and any time out is counted as 1\,200\,000 ms in our calculations.

\begin{table}
    \centering
    \caption{Automata size for the 3 robots instance showing the number of appearances of each atom.}
    \label{tbl:eval:asize}

\begin{tabular}{|ll|rrr|}
    \hline

$\varphi_i$ & \texttt{predicate} & \multicolumn{1}{c}{\WFA} &  \multicolumn{1}{c}{\WFMm} &  \multicolumn{1}{c|}{\WFMs} \\
\hline
$\varphi_1$ &\texttt{state/2} &   36          &       72 &       72 \\
&\texttt{delta/2} &  162          &      234 &      216 \\
\hline
$\varphi_2$ &\texttt{state/2} &   24          &       51 &       51 \\
&\texttt{delta/2} &   60          &      390 &      471 \\
\hline
$\varphi_3$ &\texttt{state/2} &   45          & \color{red}{-} &     372 \\
&\texttt{delta/2} &  189          & \color{red}{-} &   16\,503 \\
\hline

\end{tabular}
\end{table}
 We first compare the size of the automata in Table~\ref{tbl:eval:asize} in terms of the instances of predicates \texttt{state/2} and \texttt{delta/2}.
We see that \WFA\ generates an exponentially smaller automata, a known result from the literature~\cite{vardi95a}.
More precisely, for $\varphi_3$ the number of transitions in \WFMs\ is 90 times larger than for \WFA.
Furthermore, for this constraint, \WFMm\ reached the limit of nodes for MONA's BDD-based architecture, thus producing no result.
This outcome is based on the fact that the MSO formulas computed by \WFMm\ are significantly larger than those of \WFMs.

\begin{table}
    \centering
    \caption{Pre-processing time in milliseconds shown as $t_1/t_2$ were $t_1$ is the time for the first horizon and $t_2$ the average over subsequent calls.}
    \label{tbl:eval:ptime}
\begin{tabular}{|ll|rrrr|r|}
    \hline

$\varphi_i$ & \#r &\multicolumn{1}{c}{\WFA}& \multicolumn{1}{c}{\WFMm}& \multicolumn{1}{c}{\WFMs}&\multicolumn{1}{c}{\WFT}&\multicolumn{1}{|c|}{NC}\\
\hline
$\varphi_1$ &{2} &    \textbf{1\,194}/637    &       5\,412/638 &       5\,867/604& 2\,696/2\,992& 306/598 \\
&{3} &   \textbf{1\,991}/600  &       6\,280/671 & 6\,978/610 &     3\,390/3\,691 & 302/617 \\
\hline
$\varphi_2$ & {2} &    2\,182/579    &       33\,091/661 &       4\,966/598 & \textbf{2\,107}/2\,814 & 285/577\\
&{3} &    \textbf{1\,632}/608    & 45\,303/665 &      4\,973/604& 2\,718/3\,179 & 318/631 \\
\hline
$\varphi_3$ &{2} &    \textbf{2\,533}/599  &        \color{red}{-} &        12\,682/766 & 3\,343/3\,280 & 261/605\\
&{3} &    \textbf{3\,112}/600  & \color{red}{-} &       11,001/795 & 3,278/3,718 & 272/598\\
\hline

\end{tabular}
\end{table}

 Next, we give the preprocessing times obtained for the respective translations in Table~\ref{tbl:eval:ptime}.
For the automata-based approaches \WFA, \WFMm\ and \WFMs\ the translation is only performed once and reused in subsequent calls,
whereas for \WFT\ the translation is redone for each horizon.
The best performing approach is \WFA, for the subsequent calls the times were very similar with the exception of \WFT.
We see how for $\varphi_2$ the \WFMm\ translation takes considerably longer than for \WFMs.

\begin{table}
\centering
\caption{Statistics computed by calculating the geometric mean of all horizons.}
\label{tbl:eval:summary}
\begin{tabular}{|lll|rrrr|r|}
\hline
            {} & $\varphi_i$ & \#r &            $\WFA$ &          $\WFMm$ &    $\WFMs$ &            $\WFT$ &  $\WFNC$ \\

\hline\textbf{clingo time} &   $\varphi_1$ &  2 &             3\,374 &   \textbf{2\,788} &      2\,975 &             3\,033 &  21\,823 \\
                          &               &  3 &   \textbf{23\,173} &           27\,866 &     27\,505 &            23\,748 & 249\,737 \\
                          &   $\varphi_2$ &  2 &            10\,840 &            9\,424 &      9\,484 &    \textbf{9\,347} &  21\,378 \\
                          &               &  3 &            70\,709 &  \textbf{58\,739} &     83\,521 &            60\,765 & 246\,739 \\
                          &   $\varphi_3$ &  2 &            31\,986 &   \color{red}{-} &    606\,914 &   \textbf{16\,145} &  21\,548 \\
                          &               &  3 &            67\,287 &   \color{red}{-} &    657\,633 &   \textbf{48\,190} & 247\,718 \\
     &               &  3 &           274\,851 &   \color{red}{-} &  2\,743\,736 &  \textbf{264\,847} & 241\,752 \\                          
\hline\textbf{rules} &   $\varphi_1$ &  2 &   \textbf{89\,282} &           97\,396 &     97\,404 &            96\,793 &  77\,832 \\
     &               &  3 &  \textbf{172\,641} &          196\,220 &    190\,943 &           189\,637 & 147\,209 \\
     &   $\varphi_2$ &  2 &   \textbf{84\,180} &          122\,003 &    126\,634 &            90\,178 &  77\,832 \\
     &               &  3 &  \textbf{157\,525} &          214\,454 &    229\,063 &           166\,391 & 147\,209 \\
     &   $\varphi_3$ &  2 &   \textbf{94\,653} &   \color{red}{-} &  4\,413\,056 &           102\,687 &  77\,832 \\
     &               &  3 &  \textbf{173\,210} &   \color{red}{-} &  3\,360\,382 &           185\,155 & 147\,209 \\
    \hline\textbf{constraints} &   $\varphi_1$ &  2 &           146\,999 &          146\,323 &    146\,306 &  \textbf{140\,801} & 132\,370 \\
                          &               &  3 &           275\,747 &          274\,419 &    274\,382 &  \textbf{260\,675} & 241\,752 \\
                          &   $\varphi_2$ &  2 &  \textbf{138\,418} &          166\,449 &    171\,796 &           139\,909 & 132\,370 \\
                          &               &  3 &  \textbf{252\,023} &          295\,946 &    308\,204 &           254\,020 & 241\,752 \\
                          &   $\varphi_3$ &  2 &           153\,179 &   \color{red}{-} &  3\,341\,017 &  \textbf{147\,123} & 132\,370 \\
\hline
\end{tabular}
\end{table}
 The results of the final solving step in each workflow are summarized in Table~\ref{tbl:eval:summary},
showing the geometric mean over all horizons for obtaining a first solution.
First of all, we observe that the solving time is significantly lower when using dynamic constraints, no matter which approach is used.
For $\varphi_1$ and $\varphi_2$ the difference is negligible,
whereas for $\varphi_3$, \WFT\ is the fastest, followed by \WFA,
which is in turn twenty and ten times faster than \WFMs\ for 2 and 3 robots, respectively.
Furthermore, \WFMs\ times out for $\varphi_3$ with $\lambda=31$ and $\lambda\in\{30,31\}$ for 2 and 3 robots, respectively.
The size of the program before and after \clingo's preprocessing can be read off the number of ground rules and internal constraints, with \WFA\ having the smallest size of all approaches.
However, once the program is reduced the number of constraints shows a slight shift in favour of \WFT.

 \section{Discussion}\label{sec:discussion}

To the best of our knowledge, this work presents the first endeavor to represent dynamic constraints with automata in ASP.
The equivalence between temporal formulas and automata has been widely used in satisfiability checking, model checking, learning and synthesis \cite{vardi95a, vardi97a, giavar15a, rozvar07a, cammci19a}.
Furthermore, the field of planning has benefited from temporal reasoning to express goals and preferences using an underlying automaton~\cite{bafrbimc08a, giarub18a, baimci06a}.
There exists several systems that translate temporal formulas into automata:
SPOT \cite{dulefamirexu16a} and LTLf2DFA\footnote{ \href{https://github.com/whitemech/LTLf2DFA}{https://github.com/whitemech/LTLf2DFA}} for linear temporal logic;  \abstem~\cite{cabdie14a} and \stelp~\cite{cabdie11a} for temporal answer set programming.
Nonetheless, there have only been a few attempts to use automata-like definitions in ASP for representing temporal and procedural knowledge inspired from GOLOG programs~\cite{sobanamc03a,ryan14a}.

We investigated different automata-based implementations of dynamic (integrity) constraints using \clingo.
Our first approach was based on alternating automata, implemented entirely in ASP through meta-programming.
For our second approach, we employed the off-the-shelf automata construction tool MONA~\cite{hejejoklparasa95a} to build deterministic automata.
To this aim, we proposed two translations from dynamic logic into monadic second-order logic.
These approaches were contrasted with the temporal ASP solver \telingo\ which directly maps dynamic constraints to logic programs.
We provided an empirical analysis demonstrating the impact of using dynamic constraints to select traces among the ones induced by an associated temporal logic program.
Our study showed that the translation using solely ASP to compute an alternating automata yielded the smallest program in the shortest time.
While this approach scaled well for more complex dynamic formulas,
the MONA-based implementation performed poorly and could not handle one of our translations into second order formulas.
The best overall performance was exhibited by \telingo\ with the fundamental downside of having to redo the translation for each horizon.

Our future work aims to extend our framework to arbitrary dynamic formulas in \DELf.
Additionally, the automaton's independence of time stamps points to its potential to detect unsatisfiability
and to guide an incremental solving process.
Finally, we also intend to take advantage of \clingo's application programming interface to extend the model-ground-solve workflow of ASP with automata techniques.


\newpage
\appendix
\section{Proofs}\label{proofs}
\begin{proof}{Theorem~\ref{theorem_ldl2stm}}\label{proof_ldl2stm} 

	By double induction on $\varphi$ and $\rho$.
	
	\begin{itemize}
		\item If $\varphi = p$, with $p$ a propositional variable, $\srtm(\fovar{x},p) = \fovar{P}(\fovar{x})$. 
		If $\T, k \models p$ then $p \in T_k$. 
		By construction of $v_2$, $k \in v_2(\fovar{P})$ so $\T, v_1[\fovar{x}:=k], v_2 \models \fovar{P}(\fovar{x})$.				
		Conversely, if $\T, v_1[\fovar{x}:=k],v_2 \models \fovar{P}(\fovar{x})$ then $v_1(\fovar{x}) = k \in v_2(\fovar{P})$.
		By the construction of $v_2$, $p \in T_k$ so $\T, k \models p$.		
		\item The cases $\top$ and $\bot$ are straightforward.
		\item Negation, disjunction, conjunction and implication are proved directly by using the induction hypothesis.
		
		\item If $\varphi = \dalways{\rho}\psi$, from left to right, assume by contradiction that $\T, k \models \varphi$ but $\T, v_1[\fovar{x}:=k],v_2 \not \models \srtm(\fovar{x},\varphi)$. 
			This means that $\T, v_1[\fovar{x}:=k,\fovar{y}:=d], v_2 \models \srtp(\fovar{x}\fovar{y},\rho)$ and $\T, v_1[\fovar{x}:=k,\fovar{y}:=d],v_2 \not \models \srtm(\fovar{y},\psi)$ for some $0 \le d < \lambda$.
		By induction hypothesis, $(k,d)\in \Rel{\rho}{\T}$ and $\T, d \not \models \psi$: a contradiction.
		Conversely, assume that $\T,k \not \models \dalways{\rho}\psi$. This means that $(k,d) \in \Rel{\rho}{\T}$ and $\T,d \not \models \psi$.
		By induction, $\T, v_1[\fovar{x}:=k,\fovar{y}:=d],v_2 \models  \srtp(\fovar{x}\fovar{y},\rho)$ and $\T, v_1[\fovar{x}:=k,\fovar{y}:=d], v_2, \not \models \srtm(\fovar{y},\psi)$.
		Therefore, $\T, v_1[\fovar{x}:=k, \fovar{y}:=d], v_2 \not \models  \left(\srtp(\fovar{x}\fovar{y},\rho) \rightarrow  \srtm(\fovar{y},\psi) \right)$		
			so, $\T, v_1[\fovar{x}:=k], v_2 \not \models  \forall \fovar{y}\; \left(\srtp(\fovar{x}\fovar{y},\rho) \rightarrow  \srtm(\fovar{y},\psi) \right)$: a contradiction.

		\item If $\varphi = \deventually{\rho}\psi$ then, from left to right, if $\T, k \models \deventually{\rho}\varphi$, there exists $0 \le d< \lambda$ such that $(k,d) \in \Rel{\rho}{\T}$ and $\T,d \models \psi$.
		By induction,  $\T, v_1[\fovar{x}:=k,\fovar{y}:=d],v_2 \models  \srtp(\fovar{x}\fovar{y},\rho)$ and $\T, v_1[\fovar{y}:=d],v_2 \models \srtm(\fovar{y},\psi)$.
		Therefore, $\T, v_1[\fovar{x}:=k], v_2\models \exists \fovar{y}\; \left(\srtp(\fovar{x}\fovar{y},\rho) \wedge \srtm(\fovar{y},\psi)\right)$.
		Conversely, if $\T, v_1[\fovar{x}:=k],v_2 \models \srtm(\fovar{x},\varphi)$,  then $\T, v_1[\fovar{x}:=k,\fovar{y} :=d],v_2 \models \srtp(\fovar{x}\fovar{y},\rho)$ and $\T, v_1[\fovar{x}:=k,\fovar{y}:=d], v_2 \models \srtm(\fovar{y},\psi)$.
		By induction $(k,d) \in \Rel{\rho}{\T}$ and $\T, d \models \psi$.
		Therefore, $\T, k \models \deventually{\rho}\psi$.		
	\end{itemize}
Let us consider now the case of path formulas.
	\begin{itemize}
		\item If $\rho = \stp$, from left to right, if $(k,d) \in \Rel{\stp}{\T}$ then $d = k+1$. 
		By construction, $\T, v_1[\fovar{x}:=k, \fovar{y}:=d],v_2 \models \fovar{y} = \fovar{x}+1$.
		Conversely, if $\T, v_1[\fovar{x}:=k, \fovar{y}:=d],v_2 \models \fovar{y} = \fovar{x}+1$ then $d = k+1$ holds.
		Consequently, $(k,d) \in \Rel{\stp}{\T}$.
		\item If $\rho = \psi?$ then $\srtp(\fovar{x}\fovar{y},\rho)  = (\fovar{x} = \fovar{y} ) \wedge \srtm(\fovar{y},\psi)$.
		It holds that 
		$(k,d) \in \Rel{\rho}{\T}$ iff $k = d$ and $\T, d \models \psi$ 
		iff $k = d$ and $\T, v_1[x:=k,y:=d] ,v_2, \models \srtm(y,\psi)$ (by induction) 
			iff $\T,v_1[\fovar{x}:=k, \fovar{y}:=d] \models \srtp(\fovar{x}\fovar{y},\rho)$.
		\item If $\rho = \rho_1 + \rho_2$ then $\srtp(\fovar{x}\fovar{y}\rho) = \srtp(\fovar{x}\fovar{y},\rho_1) \vee \srtp(\fovar{x}\fovar{y},\rho_2)$.
		It holds that 
		$(k,d) \in \Rel{\rho_1 + \rho_2}{\T}$ iff either $(k,d) \in \Rel{\rho_1}{\T}$ or $(k,d) \in \Rel{\rho_2}{\T}$ 
		iff $\T, v_1[\fovar{x}:=k,\fovar{y}:=d],v_2 \models \srtp(\fovar{x}\fovar{y},\rho_1)$ or $\T, v_1[\fovar{x}:=k, \fovar{y}:=d],v_2 \models \srtp(\fovar{x}\fovar{y},\rho_2)$ (by induction) 
		iff $\T, v_1[\fovar{x}:=k,\fovar{y}:=d],v_2 \models \srtp(\fovar{x}\fovar{y},\rho)$.
	\item If $\rho = \rho_1 ; \rho_2$ then $\srtp(\fovar{x}\fovar{y},\rho) =  \exists \fovar{u} (\srtp(\fovar{x}\fovar{u},\rho_1) \wedge \srtp(\fovar{u}\fovar{y},\rho_2)) $.
		It holds that $(k,d) \in \Rel{\rho_1;\rho_2}{\T}$ iff there exists $d'$ such that  $(w,d') \in \Rel{\rho_1}{\T}$ and $(d',v) \in \Rel{\rho_2}{\T}$ 
			iff $\T,v_1[\fovar{x}:=k,\fovar{u}:=d'],v_2 \models \srtp(\fovar{x}\fovar{u},\rho_1)$ and $\T, v_1[\fovar{u}:=d', \fovar{y}:=d], v_2 \models \srtp(\fovar{u}\fovar{y},\rho_2)$ (by induction) 
			iff $\T,v_1[\fovar{x}:=k, \fovar{y}:=d] \models \srtp(\fovar{x}\fovar{y},\rho)$. 
		\item If $\rho = \rho^*$ 
		then,  from left to right, we will prove that for all $n \ge 0$ and for all $(k,d) \in \mathbb{N}\times \mathbb{N}$ if $(k,d) \in \Rel{\rho^n}{\T}$  then 
			$\T, v_1[\fovar{x}:=k, \fovar{y}:=d],v_2\models  \srtp(\fovar{x}\fovar{y},\rho^*)$ by induction on $n$
		\begin{itemize}
			\item If $n = 0$ then $k=d$. 
			Let $v_2' = v_2[\fovar{X}:=\lbrace k \rbrace]$.
			Clearly, $\T, v_1[\fovar{x}:=k, \fovar{y}:=d], v_2' \models  \fovar{X}(\fovar{x})$ and $\T, v_1[\fovar{x}:=k, \fovar{y}:=d], v_2' \models \mathtt{bound}(\fovar{X},\fovar{x},\fovar{x})$ and $\T, v_1[\fovar{x}:=k, \fovar{y}:=d], v_2' \models \mathtt{regular}(\fovar{X})$ since  $\T, v_1[\fovar{x}:=k, \fovar{y}:=d], v_2' \not \models \mathtt{succ}(\fovar{x},\fovar{x})$.
				Thanks to the second-order semantics we conclude $\T, v_1[\fovar{x}:=k, \fovar{y}:=d], v_2 \models \srtp(\fovar{x}\fovar{y},\rho^*)$.

			\item Assume that the claim holds for all $n \ge 0$ and let us prove it for $n+1$. 
			If $(k,d) \in \Rel{\rho^{n+1}}{\T}$ then, by definition, $(k,u) \in \Rel{\rho}{\T}$ and $(u,d) \in \Rel{\rho^n}{\T}$ for some $0 \le u < \lambda$.
			By Proposition~\ref{prop:ordering}, $0 \le k \le u \le d < \lambda$.
			By induction on $\rho$  we get $ \T, v_1[\fovar{x}:=k, \fovar{y}:=u], v_2 \models \srtp(\fovar{x}\fovar{y},\rho)$.
			By induction on $n$ we get $ \T, v_1[\fovar{x}:=u, \fovar{y}:=d], v_2 \models \srtp(\fovar{x}\fovar{y},\rho^*)$.
				From the previous result it follows $ \T, v_1[\fovar{x}:=u, \fovar{y}:=d], v_2' \models \fovar{X}(\fovar{x}) \wedge \fovar{X}(\fovar{y}) \wedge \mathtt{bound}(\fovar{X}, \fovar{x},\fovar{y}) \wedge \mathtt{regular}(\fovar{X})$, where $v_2'$ is an extension of $v_2$ for which $v_2'(\fovar{X})$ is defined.

			Let $v_2''$ be an extension of $v_2$ such that $v_2''(\fovar{X}) \eqdef v_2'(\fovar{X}) \cup \lbrace k \rbrace $ and $v_2'' = v_2$ for the rest of second-order variables.
			Notice that $ \T, v_1[\fovar{x}:=k, \fovar{y}:=d], v_2'' \models \fovar{X}(\fovar{x})$ and $\T, v_1[\fovar{x}:=k, \fovar{y}:=d], v_2'' \models \fovar{X}(\fovar{y})$.
			From $k \le u$ and the definition of $v_2''(\fovar{X})$, we get that $\T, v_1[\fovar{x}:=k, \fovar{y}:=d], v_2'' \models \mathtt{bound}(\fovar{X},\fovar{x},\fovar{y})$. 
			To prove that $\T, v_1[\fovar{x}:=k, \fovar{y}:=d], v_2'' \models \mathtt{regular}(\fovar{X})$, let us take $d_1, d_2 \in v_2''(\fovar{X})$. 
			We consider three cases:
				\begin{itemize}
					\item If $d_1, d_2 \in v_2'(\fovar{X})$ we use the fact that $\T, v_1[\fovar{x}:=u, y:=d], v_2' \models \mathtt{regular}(\fovar{X})$ to conclude that 
					$\T, v_1[\fovar{x}:=u, \fovar{y}:=d, \fovar{a}:=d_1, \fovar{b}:=d_2], v_2'' \models \left(\mathtt{succ}(\fovar{a},\fovar{b}) \wedge \fovar{X}(\fovar{a}) \wedge \fovar{X}(\fovar{b})) \to \srtp(\fovar{a}\fovar{b},\rho)\right)$ 					
				\item $d_1 = d_2 = k$ then  $\T, v_1[\fovar{x}:=u, \fovar{y}:=d, \fovar{a}:=d_1, \fovar{b}:=d_2], v_2'' \models \left(\mathtt{succ}(\fovar{a},\fovar{b}) \wedge \fovar{X}(\fovar{a}) \wedge \fovar{X}(\fovar{b})) \to \srtp(\fovar{a}\fovar{b},\rho)\right)$  since  \newline $\T, v_1[\fovar{x}:=u, \fovar{y}:=d, \fovar{a}:=d_1, \fovar{b}:=d_2], v_2'' \not \models (\mathtt{succ}(\fovar{a},\fovar{b})$.
				\item $d_1 = k$ and $d_2 \not = w$. Necessarily, $d_1 = u$.  In this case, \newline $\T, v_1[\fovar{x}:=u, \fovar{y}:=d, \fovar{a}:=d_1, \fovar{b}:=d_2], v_2'' \models \newline \left(\mathtt{succ}(\fovar{a},\fovar{b}) \wedge \fovar{X}(\fovar{a}) \wedge \fovar{X}(\fovar{b})) \to \srtp(\fovar{a}\fovar{b},\rho)\right)$ 
				\newline because $\T, v_1[\fovar{x}:=u, \fovar{y}:=d, \fovar{a}:=d_1, \fovar{b}:=d_2], v_2'' \models \srtp(\fovar{a}\fovar{b},\rho)$.
				\end{itemize}
		Thus, we conclude $\T, v_1[\fovar{x}:=k, \fovar{y}:=d], v_2'' \models \mathtt{regular}(\fovar{X})$.
		\end{itemize}
		
			for the converse direction, if $\T, v_1[\fovar{x}:=k, \fovar{y}:=d],v_2 \models  \srtp(\fovar{x}\fovar{y},\rho^*)$ then there exists an assignment $v_2'$ that extends $v_2$ with an assignment for the second-order variable $\fovar{X}$. By definition, it holds that 

		\begin{enumerate}
			\item $\T, v_1[\fovar{x}:=k, \fovar{y}:=d],v_2' \models \fovar{X}(\fovar{x})$
			\item $\T, v_1[\fovar{x}:=k, \fovar{y}:=d],v_2' \models \fovar{X}(\fovar{y})$
			\item $\T, v_1[\fovar{x}:=k, \fovar{y}:=d],v_2' \models \mathtt{bound}( \fovar{X},\fovar{x},\fovar{y})$
			\item $\T, v_1[\fovar{x}:=k, \fovar{y}:=d],v_2' \models \mathtt{regular}(\fovar{X})$
		\end{enumerate}
			\noindent From all those items we get that there exists $u_0, u_1, \cdots, u_n$ in $v_2(\fovar{X})$ such that $u_0 = k$, $u_n = d$ and for all $0 \le i < n$, $(u_i, u_{i+1}) \in \Rel{\rho}{\T}$. By using the definition of $\Rel{\rho^*}{\T}$, we conclude that $(k,d) \in \Rel{\rho^*}{\T}$.				
\end{itemize}
\end{proof}

 \begin{proof}{Theorem~\ref{theorem_ldl2mso}}\label{proof_ldl2mso}
	
	If $\varphi$ is propositional atom $p$ then $\mso(\fovar{t}, p) = P(\fovar{t})$. It is true that $\T, k \models \varphi$ iff $\T, v_1[\fovar{t}:=k],v_2\models \mso(\fovar{t},p)$. 
If $\varphi$ is a nonatomic formula, we prove this theorem in two directions. 
	
	Suppose first that $\T,k \models \varphi$.
	Let $v_2'$ be a second-order assignment such that $v_2'(\fovar{Q_{\theta_i}}) = \lbrace \lbrace x \mid \T, x \models \Theta_i\rbrace\rbrace$ for each $\Theta_i \in \cl{\varphi}$ and $v_2'(\fovar{P}) = v_2(\fovar{P})$ otherwise.
	By assumption, $\T, v_1[\fovar{t}:=k],v_2' \models \fovar{Q_{\varphi}}(\fovar{t})$.	
	It remains to prove that $ \T, v_1[\fovar{t}:=k], v_2' \models \forall \fovar{x}.\; t(\Theta_i,\fovar{x})$  for each nonatomic subformula $\Theta_i \in cl(\varphi)$, which we prove by induction over $\Theta_i$ 
	
	\begin{itemize}
		\item If $\Theta_i = \neg \Theta_j$, then $t(\Theta_i,\fovar{x}) = \left( \fovar{Q_{\Theta_i}} (\fovar{x}) \leftrightarrow \neg \fovar{Q_{\Theta_j}}(\fovar{x})\right)$. This holds, since $v_2'(\fovar{Q_{\neg \Theta_j}}) = \lbrace x \mid \T, x \not \models \Theta_j\rbrace$ and $v_2'(\fovar{Q_{\Theta_j}}) = \lbrace x \mid \T, x \models \Theta_j \rbrace$.
		\item If $\Theta_i = \Theta_j \wedge \Theta_k$ , then $t(\Theta_i, \fovar{x}) = \left(\fovar{Q_{\Theta_i}}(\fovar{x}) \leftrightarrow \left( \fovar{Q_{\Theta_j}}(\fovar{x}) \wedge \fovar{Q_{\Theta_j}}(\fovar{x}) \right)\right)$. This holds since $v_2'(\fovar{Q_{\Theta_j \wedge \Theta_k}}) = \lbrace  x \mid \T, x \models \Theta_j \hbox{ and } \T,x \models \Theta_k\rbrace$, $v_2'(\fovar{Q_{\Theta_j}}) = \lbrace  x \mid \T, x \models \Theta_j \rbrace$ and $v_2'(\fovar{Q_{\Theta_k}}) = \lbrace  x \mid \T, x \models \Theta_k\rbrace$.
		\item If $\Theta_i = \Theta_j \vee \Theta_k$ , then $t(\Theta_i, \fovar{x}) = \left(\fovar{Q_{\Theta_i}}(\fovar{x}) \leftrightarrow \left( \fovar{Q_{\Theta_j}}(\fovar{x}) \vee \fovar{Q_{\Theta_j}}(\fovar{x}) \right)\right)$. This holds since $v_2'(\fovar{Q_{\Theta_j \vee \Theta_k}}) = \lbrace  x \mid \T, x \models \Theta_j \hbox{ or } \T,x \models \Theta_k\rbrace$, $v_2'(\fovar{Q_{\Theta_j}}) = \lbrace  x \mid \T, x \models \Theta_j \rbrace$ and $v_2'(\fovar{Q_{\Theta_k}}) = \lbrace  x \mid \T, x \models \Theta_k\rbrace$.
		\item If $\Theta_i = \Theta_j \rightarrow \Theta_k$ , then $t(\Theta_i, \fovar{x}) = \left(\fovar{Q_{\Theta_i}}(\fovar{x}) \leftrightarrow \left( \fovar{Q_{\Theta_j}}(\fovar{x}) \rightarrow \fovar{Q_{\Theta_j}}(\fovar{x}) \right)\right)$. This holds since $v_2'(\fovar{Q_{\Theta_j \rightarrow \Theta_k}}) = \lbrace  x \mid \T, x \not \models \Theta_j \hbox{ or } \T,x \models \Theta_k\rbrace$, $v_2'(\fovar{Q_{\Theta_j}}) = \lbrace  x \mid \T, x \models \Theta_j \rbrace$ and $v_2'(\fovar{Q_{\Theta_k}}) = \lbrace  x \mid \T, x \models \Theta_k\rbrace$.
		\item If $\Theta_i = \dalways{\Theta_j?}\Theta_k$ we proceed as in the case of implication.
		\item If $\Theta_i = \deventually{\Theta_j?}\Theta_k$ we proceed as in the case of conjunction. 
		\item If $\Theta_i = \dalways{\stp}\Theta_j$, then $t(\Theta_i,\fovar{x}) = \fovar{Q_{\Theta_i}}(\fovar{x}) \leftrightarrow \left(\forall \fovar{y}\; \left(\fovar{y} = \fovar{x}+1\right) \rightarrow \fovar{Q_{\Theta_j}} (\fovar{y})\right)$. This holds since 	
		\begin{eqnarray*}
			v_2'(\fovar{Q_{\Theta_i}})  &=& \lbrace x \mid \T, x \models \dalways{\stp} \Theta_k \rbrace \\
			& =&\lbrace  x \mid \hbox{ for all } (x,y) \hbox{ if } (x,y) \in \Rel{\stp}{\T} \hbox{ then }  \T, y \models \Theta_j \rbrace\\
			& =&\lbrace  x \mid  \hbox{ for all y, if} y = x+1 \hbox{ then }  \T, y \models \Theta_j \rbrace.
		\end{eqnarray*}
	
			\noindent and $v_2'(\fovar{Q_{\Theta_j}}) = \lbrace x \mid \T, x \models \Theta_j \rbrace$.
		\item If $\Theta_i = \deventually{\stp}\Theta_j$, then $t(\Theta_i,\fovar{x}) = \fovar{Q_{\Theta_i}}(\fovar{x}) \leftrightarrow \left(  \exists \fovar{y}\; \left(\fovar{y} = \fovar{x}+1\right) \rightarrow \fovar{Q_{\Theta_j}} (\fovar{y})\right)$. This holds because 
		\begin{eqnarray*}
			v_2'(\fovar{Q_{\Theta_i}})  &=& \lbrace x \mid \T, x \models \deventually{\stp}\Theta_j \rbrace \\
						  &=& \lbrace x \mid \hbox{ there exits }  (x,y) \in \Rel{\stp}{\T} \hbox{ such that }\T, y \models \Theta_j \rbrace \\
						  &=& \lbrace x \mid \hbox{ there exits }  y = x+1  \hbox{ such that }\T, y \models \Theta_j \rbrace \\
		\end{eqnarray*}
		\noindent and $v_2'(\fovar{Q_{\Theta_j}}) = \lbrace x \mid \T, x \models \Theta_j \rbrace$.

	\item If $\Theta_i = \dalways{\rho_1;\rho_2}\Theta_j$, then $t(\Theta_i,\fovar{x}) = \fovar{Q_{\Theta_i}}(\fovar{x}) \leftrightarrow \left(\fovar{Q_{\dalways{\rho_1}\dalways{\rho_2} \Theta_j }}(\fovar{x})\right)$. 
		This holds because $v_2'(\fovar{Q_{\Theta_i}})  = \lbrace x \mid \T, x \models \dalways{\rho_1;\rho_2}\Theta_j\rbrace  = \lbrace x \mid \T, x \models \dalways{\rho_1}\dalways{\rho_2}\Theta_j\rbrace$ 
			by Proposition~\ref{prop:validities} (item 3) and $v_2'(\fovar{Q_{\dalways{\rho_1}\dalways{\rho_2}\Theta_j}}) = \lbrace x \mid \T, x \models \dalways{\rho_1}\dalways{\rho_2}\Theta_j\rbrace$.	
		\item If $\Theta_i = \deventually{\rho_1;\rho_2}\Theta_j$, then $t(\Theta_i,\fovar{x}) = \fovar{Q_{\Theta_i}}(x) \leftrightarrow \left(\fovar{Q_{\deventually{\rho_1}\deventually{\rho_2} \Theta_j}} (\fovar{x}) \right)$. 
			This holds because $v_2'(\fovar{Q_{\Theta_i}})  = \lbrace x \mid \T, x \models \deventually{\rho_1;\rho_2}\Theta_j\rbrace  = \lbrace x \mid \T, x \models \deventually{\rho_1}\deventually{\rho_2}\Theta_j\rbrace$  thanks to Proposition~\ref{prop:validities} (item 4) and $v_2'(\fovar{Q_{\deventually{\rho_1}\deventually{\rho_2}\Theta_j}}) = \lbrace x \mid \T, x \models \deventually{\rho_1}\deventually{\rho_2}\Theta_j\rbrace$.
			
		\item If $\Theta_i = \dalways{\rho_1+\rho_2}\Theta_j$, then $t(\Theta_i,\fovar{x}) = \fovar{Q_{\Theta_i}}(\fovar{x}) \leftrightarrow \left(\fovar{Q_{\dalways{\rho_1} \Theta_j}}(\fovar{x}) \wedge \fovar{Q_{\dalways{\rho_2} \Theta_j}} (\fovar{x}) \right)$. 
			This holds because $v_2'(\fovar{Q_{\Theta_i}})  = \lbrace x \mid \T, x \models \dalways{\rho_1 + \rho_2}\Theta_j\rbrace  = \lbrace x \mid \T, x \models \dalways{\rho_1} \Theta _j \wedge \dalways{\rho_2}\Theta_j\rbrace$ by Proposition~\ref{prop:validities} (item 1), $v_2'(\fovar{Q_{\dalways{\rho_1}\Theta_j}}) = \lbrace x \mid \T, x \models \dalways{\rho_1}\Theta_j\rbrace$ and $v_2'(\fovar{Q_{\dalways{\rho_2}\Theta_j}}) = \lbrace x \mid \T, x \models \dalways{\rho_2}\Theta_j\rbrace$.
	
		\item If $\Theta_i = \deventually{\rho_1+\rho_2}\Theta_j$, then $t(\Theta_i,\fovar{x}) = \fovar{Q_{\Theta_i}(x)} \leftrightarrow \left(\fovar{Q_{\deventually{\rho_1}\Theta_j}}(\fovar{x}) \vee \fovar{Q_{\deventually{\rho_2} \Theta_j}}(\fovar{x})\right)$.
			This holds because $v_2'(\fovar{Q_{\Theta_i}})  = \lbrace x \mid \T, x \models \deventually{\rho_1 + \rho_2}\Theta_j\rbrace  = \lbrace x \mid \T, x \models \deventually{\rho_1} \Theta _j \vee \deventually{\rho_2}\Theta_j\rbrace$ by Proposition~\ref{prop:validities} (item 2), $v_2'(\fovar{Q_{\deventually{\rho_1}\Theta_j}}) = \lbrace x \mid \T, x \models \deventually{\rho_1}\Theta_j\rbrace$ and $v_2'(\fovar{Q_{\deventually{\rho_2}\Theta_j}}) = \lbrace x \mid \T, x \models \deventually{\rho_2}\Theta_j\rbrace$.
	
		\item If $\Theta_i = \dalways{\rho^*}\Theta_j$, then $t(\Theta_i,\fovar{x}) = \fovar{Q_{\Theta_i}(x)} \leftrightarrow \left(\fovar{Q_{\Theta_j}} (\fovar{x}) \wedge \fovar{Q_{\dalways{\rho} \dalways{\rho^*}\Theta_j}}(\fovar{x})\right)$. This holds since $v_2'(\fovar{Q_{\Theta_i}})  = \lbrace x \mid \T, x \models \dalways{\rho^*}\Theta_j\rbrace  = \lbrace x \mid \T, x \models \Theta_j \wedge \dalways{\rho}\dalways{\rho^*} \Theta_j \rbrace$ by Proposition~\ref{prop:validities} (item 5), $v_2'(\fovar{Q_{\Theta_j}}) = \lbrace x \mid \T, x \models \Theta_j\rbrace$ and $v_2'(\fovar{Q_{\dalways{\rho}\dalways{\rho^*}\Theta_j}}) = \lbrace x \mid \T, x \models \dalways{\rho}\dalways{\rho^*}\Theta_j\rbrace$.
	
		\item If $\Theta_i = \deventually{\rho^*}\Theta_j$, then $t(\Theta_i,x) = \fovar{Q_{\Theta_i}}(\fovar{x}) \leftrightarrow \left( \fovar{Q_{\Theta_j}}(\fovar{x}) \vee \fovar{Q_{\deventually{\rho} \deventually{\rho^*}\Theta_j}}(\fovar{x})\right)$. This holds since $v_2'(\fovar{Q_{\Theta_i}})  = \lbrace x \mid \T, x \models \deventually{\rho^*}\Theta_j\rbrace  = \lbrace x \mid \T, x \models \Theta_j \vee \deventually{\rho}\deventually{\rho^*} \Theta_j \rbrace$ by Proposition~\ref{prop:validities} (item 6), $v_2'(\fovar{Q_{\Theta_j}}) = \lbrace x \mid \T, x \models \Theta_j\rbrace$ and $v_2'(\fovar{Q_{\deventually{\rho}\deventually{\rho^*}\Theta_j}}) = \lbrace x \mid \T, x \models \deventually{\rho}\deventually{\rho^*}\Theta_j\rbrace$.
	\end{itemize}
	
	Assume now that $\T, v_1[\fovar{t}:=k], v_2 \models \mso(\fovar{t},\varphi)$. 
	This means that there is an assignment $v_2'$ that extends $v_2$ by defining $v_2'(\fovar{Q_{\theta_i}})$ for each predicate $\fovar{Q_{\theta_i}}$  with $\theta_i$ being a non-atomic formula  in $\cl{\varphi}$ and satisfying 
	 $\T, v_1[\fovar{t}:=k], v_2' \models \fovar{Q_{\varphi}}(\fovar{t}) \wedge (\forall \fovar{x} (\land_{i=0}^m t(\theta_i,\fovar{x}) )$.
	We now prove by induction on $\varphi$ that if $\T, v_1[\fovar{t}:=k, \fovar{x}:=d],v_2' \models  \fovar{Q_{\varphi}}(\fovar{x})$ then $\T, d \models \varphi$ for all $0\le d < \lambda$ so 
	$\T, v_1[\fovar{t}:=k, \fovar{x}:=k], v_2' \models  \fovar{Q_{\varphi}(x)}$ indicates that $\T, k \models \varphi$.
	
	\begin{itemize}
		\item If $\varphi = \neg \Theta_j$, then $t(\varphi,\fovar{x}) = \left( \fovar{Q_\varphi}(\fovar{x}) \leftrightarrow \neg \fovar{Q_{\Theta_j}}(\fovar{x})\right)$.
		Since $\T, v_1[\fovar{t}:=k, \fovar{x}:=d],v_2' \models  t(\varphi,\fovar{x})$ for all $0 \le d < \lambda$,		
		it holds that $d \in v_2'(\fovar{Q_\varphi})$ iff $d \not \in v_2'(\fovar{Q_{\Theta_j}})$. 
		By induction we get $\T, d \not \models \Theta_j$. Thus, $\T,d \models \varphi$.
	\item If $\varphi = \Theta_j \wedge \Theta_k$ then $t(\varphi,\fovar{x}) = \left( \fovar{Q_\varphi}(\fovar{x}) \leftrightarrow \fovar{Q_{\Theta_j}}(\fovar{x}) \wedge \fovar{Q_{\Theta_k}}(\fovar{x})\right)$. Since 
		$\T, v_1[\fovar{t}:=k, \fovar{x}:=d],v_2' \models  t(\varphi,\fovar{x})$ for all $0 \le d < \lambda$,		
		it follows that $d \in v_2'(\fovar{Q_\varphi})$ iff $d \in v_2'(\fovar{Q_{\Theta_k}}) \cap v_2'(\fovar{Q_{\Theta_j}})$. 						
			By induction $\T, d \models \Theta_j$ and $\T,d \models \Theta_k$.
		\item If $\varphi = \Theta_j \vee \Theta_k$ then $t(\varphi,\fovar{x}) = \left( \fovar{Q_\varphi}(\fovar{x}) \leftrightarrow \fovar{Q_{\Theta_j}}(\fovar{x}) \vee \fovar{Q_{\Theta_k}}(\fovar{x})\right)$. Since
		$\T, v_1[\fovar{t}:=k, \fovar{x}:=d],v_2' \models  t(\varphi,\fovar{x})$ for all $0 \le d < \lambda$,		
		it follows that $d \in v_2'(\fovar{Q_\varphi})$ iff $d \in v_2'(\fovar{Q_{\Theta_k}}) \cup v_2'(\fovar{Q_{\Theta_j}})$. By induction $\T, d \models \Theta_j$ or $\T,d \models \Theta_k$.

	\item If $\varphi = \Theta_j \rightarrow \Theta_k$ then $t(\varphi,\fovar{x}) = \left( \fovar{Q_\varphi}(\fovar{x}) \leftrightarrow \left( \fovar{Q_{\Theta_j}}(\fovar{x}) \rightarrow \fovar{Q_{\Theta_k}}(\fovar{x})\right)\right)$. 		Since $\T, v_1[\fovar{t}:=k, \fovar{x}:=d],v_2' \models  t(\varphi,\fovar{x})$ for all $0 \le d < \lambda$,
	it follows that $v_1(\fovar{x}) \in v_2'(\fovar{Q_\varphi})$ iff $d \in \overline{v_2(\fovar{Q_{\Theta_k}})} \cup v_2'(\fovar{Q_{\Theta_j}})$. By induction $\T, d \not \models \Theta_j$ and $\T,d \models \Theta_k$ meaning that $\T,d \models \Theta_j \rightarrow \Theta_k$.

		\item If $\varphi = \dalways{\Theta_k ? } \Theta_j$ we proceed as for implication.
		\item If $\varphi = \deventually{\Theta_k ? } \Theta_j$ we proceed as for conjunction.
		\item If $\varphi = \dalways{\stp} \Theta_j$ then $t(\varphi,\fovar{x}) = \left( Q_\varphi(x) \leftrightarrow \left( \forall \fovar{y}.; \fovar{y} = \fovar{x}+1 \rightarrow  \fovar{Q_{\Theta_j}}(\fovar{y})\right)\right)$. 
		Since $\T, v_1[\fovar{t}:=k, \fovar{x}:=d],v_2' \models  t(\varphi,\fovar{x})$ for all $0 \le d < \lambda$,	
		it follows that $d \in v_2'(\fovar{Q_\varphi})$ iff for all $\fovar{y}$, $\fovar{y}= \fovar{d}+1$ implies $\fovar{y} \in v_2'(\fovar{Q_{\Theta_j}})$. 
		By induction it follows that either $d +1 = \lambda$ or $\T, d+1 \models \Theta_j$ so $\T, d \models \varphi$.
	\item If $\varphi = \deventually{\stp} \Theta_j$ then $t(\varphi,\fovar{x}) = \left( \fovar{Q_\varphi}(\fovar{x}) \leftrightarrow \left( \exists \fovar{y}\; \left(\fovar{y} = \fovar{x}+1 \wedge  \fovar{Q_{\Theta_j}}(\fovar{y})\right)\right)\right)$. 
		Since $\T, v_1[\fovar{t}:=k, \fovar{x}:=d],v_2' \models  t(\varphi,\fovar{x})$ for all $0 \le d < \lambda$,		
		it follows that $d \in v_2'(\fovar{Q_\varphi})$ iff there exists $\fovar{y} = d+1$ and $\fovar{y} \in v_2'(\fovar{Q_{\Theta_j}})$. 
		By induction it follows that $d+1 < \lambda$ and $\T, d+1 \models \fovar{\Theta_j}$ so $\T, d \models \varphi$.
		
		\item If $\varphi = \dalways{\rho_1;\rho_2} \Theta_j$ then $t(\varphi,\fovar{x}) = \left( \fovar{Q_\varphi}(\fovar{x}) \leftrightarrow  \fovar{Q_{\dalways{\rho_1}\dalways{\rho_2} \Theta_j}}(\fovar{x})\right)$. 
		Since \newline $\T, v_1[\fovar{t}:=k, \fovar{x}:=d],v_2' \models  t(\varphi,\fovar{x})$ for all $0 \le d < \lambda$,		
		it follows that $d \in v_2'(\fovar{Q_\varphi})$ iff  $d \in v_2'(\fovar{Q_{\dalways{\rho_1}\dalways{\rho_2} \Theta_j}})$. 
			By induction it follows that there $\T, d \models \dalways{\rho_1}\dalways{\rho_2} \Theta_j$ and, by Proposition~\ref{prop:validities} (item 3) so $\T, d \models \varphi$.

		\item If $\varphi = \deventually{\rho_1;\rho_2} \Theta_j$ then $t(\varphi,\fovar{x}) = \left( \fovar{Q_\varphi}(\fovar{x}) \leftrightarrow \fovar{Q_{\deventually{\rho_1}\deventually{\rho_2} \Theta_j}}(\fovar{x})\right)$. 
		Since \newline $\T, v_1[\fovar{t}:=k, \fovar{x}:=d],v_2' \models  t(\varphi,\fovar{x})$ for all $0 \le d < \lambda$,		
		it follows that $d \in v_2'(\fovar{Q_\varphi})$ iff  $d \in v_2'(\fovar{Q_{\deventually{\rho_1}\deventually{\rho_2} \Theta_j}})$. 
		By induction it follows that there $\T, d \models \deventually{\rho_1}\deventually{\rho_2} \Theta_j$ and, by Proposition~\ref{prop:validities} (item 4) so $\T, d \models \varphi$.
	\item If $\varphi = \dalways{\rho_1 + \rho_2} \Theta_j$ then $t(\varphi,\fovar{x}) = \left( \fovar{Q_\varphi}(\fovar{x}) \leftrightarrow  \left(\fovar{Q_{\dalways{\rho_1}\Theta_j}}(\fovar{x}) \wedge \fovar{Q_{\dalways{\rho_2}\Theta_j}}(\fovar{x})\right)\right)$. 
		Since $\T, v_1[\fovar{t}:=k, \fovar{x}:=d],v_2' \models  t(\varphi,\fovar{x})$ for all $0 \le d < \lambda$,		
		it follows that $d \in v_2'(\fovar{Q_\varphi})$ iff  $d \in v_2'(\fovar{Q_{\dalways{\rho_1} \Theta_j}})\cap  v_2'(\fovar{Q_{\dalways{\rho_2} \Theta_j}})$. 
		By induction it follows that there $\T, d \models \dalways{\rho_1}\Theta_j \wedge \dalways{\rho_2}\Theta_j$ and, by Proposition~\ref{prop:validities} (item 1) so $\T, d \models \varphi$.

	\item If $\varphi = \deventually{\rho_1 + \rho_2} \Theta_j$ then $t(\varphi,\fovar{x}) = \left( \fovar{Q_\varphi}(\fovar{x}) \leftrightarrow  \left(\fovar{Q_{\deventually{\rho_1}\Theta_j}}(\fovar{x}) \vee \fovar{Q_{\deventually{\rho_2}\Theta_j}}(\fovar{x})\right)\right)$. 
		Since $\T, v_1[\fovar{t}:=k, \fovar{x}:=d],v_2' \models  t(\varphi,\fovar{x})$ for all $0 \le d < \lambda$,		
		it follows that $d \in v_2'(\fovar{Q_\varphi})$ iff  $d \in v_2'(\fovar{Q_{\deventually{\rho_1} \Theta_j}})\cup v_2'(\fovar{Q_{\deventually{\rho_2} \Theta_j}})$. 
		By induction, it follows that there $\T, d \models \deventually{\rho_1}\Theta_j \vee \deventually{\rho_2}\Theta_j$ and, by Proposition~\ref{prop:validities} (item 2) so $\T, d \models \varphi$.

	\item If $\varphi = \dalways{\rho^*} \Theta_j$ then $t(\varphi,\fovar{x}) = \left( \fovar{Q_\varphi}(\fovar{x}) \leftrightarrow  \left(\fovar{Q_{\Theta_j}} (\fovar{x}) \wedge \fovar{Q_{\dalways{\rho}\dalways{\rho^*}\Theta_j}}(\fovar{x}) \right)\right)$.
		Since $\T, v_1[\fovar{t}:=k, \fovar{x}:=d],v_2' \models  t(\varphi,\fovar{x})$ for all $0 \le d < \lambda$,		
		it follows that $d \in v_2'(\fovar{Q_\varphi})$ iff  $d \in v_2'(\fovar{Q_{\Theta_j}})\cap  v_2'(\fovar{Q_{\dalways{\rho}\dalways{\rho^*} \Theta_j}})$. 
		By induction it follows that there $\T, d \models \Theta_j \wedge \dalways{\rho}\dalways{\rho^*}\Theta_j$.
		By Proposition~\ref{prop:validities} (item 5) so $\T, d \models \varphi$.

		\item If $\varphi = \deventually{\rho^*} \Theta_j$ then $t(\varphi,x) = \left( Q_\varphi(x) \leftrightarrow  \left(Q_{\Theta_j} (x) \vee Q_{\deventually{\rho}\deventually{\rho^*}\Theta_j}(x) \right)\right)$.
		Since $\T, v_1[\fovar{t}:=k, \fovar{x}:=d],v_2' \models  t(\varphi,\fovar{x})$ for all $0 \le d < \lambda$,						
		it follows that $d \in v_2'(\fovar{Q_\varphi})$ iff  $d \in v_2'(\fovar{Q_{\Theta_j}})\cup  v_2'(\fovar{Q_{\dalways{\rho}\dalways{\rho^*} \Theta_j}})$. 
		By induction it follows that there $\T, d \models \Theta_j \vee \deventually{\rho}\deventually{\rho^*}\Theta_j$.
		By Proposition~\ref{prop:validities} (item 6) so $\T, d \models \varphi$.
	\end{itemize}	
	\end{proof}	
  \section{Detailed results tables}
In the following tables, lambdas appear crossed out when the instance was UNSAT with the corresponding constraint. The results are for finding the first model and the best performance excluding NC is found in bold.
\begin{table}[h!]
\centering
\caption{Statistics for constraint $\varphi_1$ and the 2 robots instance.   }
\label{tbl:eval:d1:r2}
\begin{tabular}{|ll|rrrr|r|}
\hline
                          &  $\lambda$ &            $\WFA$ &          $\WFMm$ &         $\WFMs$ &            $\WFT$ &  $\WFNC$ \\

    \hline\textbf{translation time} &  \sout{25} &    \textbf{1\,194} &            5\,412 &           5\,867 &             2\,696 &     305 \\
                        &  \sout{26} &               557 &     \textbf{451} &             725 &             3\,407 &     397 \\
                        &         27 &               837 &              914 &    \textbf{638} &             2\,683 &     558 \\
                        &         28 &      \textbf{589} &              648 &             617 &             3\,218 &     454 \\
                        &         29 &               508 &              670 &    \textbf{448} &             2\,873 &     799 \\
                        &         30 &               663 &     \textbf{445} &             472 &             2\,197 &     768 \\
                        &         31 &      \textbf{672} &              701 &             724 &             3\,574 &     614 \\
    \hline\textbf{clingo time} &  \sout{25} &             3\,395 &   \textbf{2\,387} &           2\,775 &             2\,657 &  10\,104 \\
                            &  \sout{26} &             3\,900 &            3\,732 &  \textbf{3\,016} &             3\,375 &  14\,131 \\
                            &         27 &             4\,497 &            2\,335 &  \textbf{2\,128} &             3\,769 &  31\,167 \\
                            &         28 &    \textbf{2\,153} &            3\,779 &           3\,509 &             2\,434 &  29\,118 \\
                            &         29 &             3\,032 &            2\,826 &  \textbf{2\,796} &             3\,280 &  25\,184 \\
                            &         30 &             2\,700 &   \textbf{2\,183} &           2\,332 &             3\,730 &  24\,176 \\
                            &         31 &             4\,744 &            2\,700 &           5\,059 &    \textbf{2\,348} &  29\,871 \\
   \hline\textbf{choices} &  \sout{25} &            15\,197 &           14\,196 &          16\,509 &   \textbf{11\,267} &  51\,711 \\
                          &  \sout{26} &            24\,236 &           21\,258 &          17\,956 &   \textbf{14\,982} &  91\,708 \\
                          &         27 &            27\,412 &  \textbf{15\,038} &          15\,995 &            19\,005 & 120\,714 \\
                          &         28 &            17\,473 &           23\,631 &          41\,320 &   \textbf{12\,952} & 124\,386 \\
                          &         29 &            25\,394 &           20\,823 &          20\,373 &   \textbf{17\,310} & 121\,999 \\
                          &         30 &            23\,611 &  \textbf{16\,924} &          18\,829 &            21\,436 & 125\,925 \\
                          &         31 &            37\,079 &  \textbf{20\,815} &          36\,333 &            21\,725 & 150\,685 \\
 \hline\textbf{conflicts} &  \sout{25} &             5\,916 &            5\,927 &           6\,933 &    \textbf{4\,986} &  31\,833 \\
                          &  \sout{26} &             9\,397 &            9\,302 &           7\,627 &    \textbf{6\,551} &  40\,969 \\
                          &         27 &            10\,591 &   \textbf{5\,608} &           5\,710 &             7\,832 &  80\,625 \\
                          &         28 &             5\,255 &            9\,671 &           8\,523 &    \textbf{4\,241} &  77\,873 \\
                          &         29 &             7\,281 &            7\,353 &           6\,550 &    \textbf{6\,029} &  71\,170 \\
                          &         30 &             6\,391 &   \textbf{5\,065} &           5\,577 &             7\,012 &  71\,750 \\
                          &         31 &            11\,411 &            6\,690 &          13\,001 &    \textbf{3\,443} &  84\,896 \\
    \hline\textbf{rules} &  \sout{25} &   \textbf{77\,980} &           85\,216 &          85\,224 &            84\,688 &  67\,749 \\
                        &  \sout{26} &   \textbf{81\,860} &           89\,396 &          89\,404 &            88\,842 &  71\,213 \\
                        &         27 &   \textbf{85\,740} &           93\,576 &          93\,584 &            92\,996 &  74\,677 \\
                        &         28 &   \textbf{89\,620} &           97\,756 &          97\,764 &            97\,150 &  78\,141 \\
                        &         29 &   \textbf{93\,500} &          101\,936 &         101\,944 &           101\,304 &  81\,605 \\
                        &         30 &   \textbf{97\,380} &          106\,116 &         106\,124 &           105\,458 &  85\,069 \\
                        &         31 &  \textbf{101\,260} &          110\,296 &         110\,304 &           109\,612 &  88\,533 \\
      \hline\textbf{constraints} &  \sout{25} &           125\,800 &          124\,948 &         124\,932 &  \textbf{120\,596} & 113\,198 \\
                          &  \sout{26} &           133\,109 &          132\,321 &         132\,305 &  \textbf{127\,561} & 119\,809 \\
                          &         27 &           140\,418 &          139\,694 &         139\,678 &  \textbf{134\,526} & 126\,420 \\
                          &         28 &           147\,727 &          147\,067 &         147\,051 &  \textbf{141\,491} & 133\,031 \\
                          &         29 &           155\,036 &          154\,440 &         154\,424 &  \textbf{148\,456} & 139\,642 \\
                          &         30 &           162\,345 &          161\,813 &         161\,797 &  \textbf{155\,421} & 146\,253 \\
                          &         31 &           169\,654 &          169\,186 &         169\,170 &  \textbf{162\,386} & 152\,864 \\
\hline
\end{tabular}
\end{table}
 \begin{table}[h!]
\centering
\caption{Statistics for constraint $\varphi_1$ and the 3 robots instance.   }
\label{tbl:eval:d1:r3}
\begin{tabular}{|ll|rrrr|r|}
\hline
                          &  $\lambda$ &            $\WFA$ &           $\WFMm$ &          $\WFMs$ &            $\WFT$ &    $\WFNC$ \\

    \hline\textbf{translation time} &  \sout{25} &    \textbf{1\,991} &             6\,280 &            6\,978 &             3\,390 &       301 \\
                        &  \sout{26} &      \textbf{474} &               477 &              689 &             4\,123 &       477 \\
                        &  \sout{27} &               670 &               937 &     \textbf{632} &             3\,316 &       611 \\
                        &  \sout{28} &      \textbf{633} &               634 &              637 &             3\,578 &       388 \\
                        &  \sout{29} &               574 &               712 &     \textbf{476} &             3\,240 &       751 \\
                        &         30 &               628 &               501 &     \textbf{490} &             3\,601 &       842 \\
                        &         31 &      \textbf{623} &               767 &              738 &             4\,289 &       629 \\
    \hline\textbf{clingo time} &  \sout{25} &            12\,069 &            11\,375 &           11\,196 &   \textbf{10\,687} &    49\,097 \\
                        &  \sout{26} &   \textbf{12\,487} &            15\,689 &           13\,500 &            13\,588 &    77\,250 \\
                        &  \sout{27} &   \textbf{16\,865} &            26\,711 &           20\,568 &            17\,957 &   193\,067 \\
                        &  \sout{28} &            38\,528 &   \textbf{28\,985} &           29\,708 &            38\,543 &   530\,237 \\
                        &  \sout{29} &   \textbf{49\,117} &            52\,888 &           60\,487 &            49\,148 &   796\,953 \\
                        &         30 &            36\,833 &            62\,928 &           55\,439 &   \textbf{36\,765} &   508\,369 \\
                        &         31 &   \textbf{20\,253} &            28\,378 &           38\,456 &            23\,454 &   385\,140 \\
   \hline\textbf{choices} &  \sout{25} &            59\,635 &            49\,477 &          117\,691 &   \textbf{45\,287} &   173\,746 \\
                          &  \sout{26} &            64\,653 &            64\,926 &           62\,331 &   \textbf{59\,828} &   245\,978 \\
                          &  \sout{27} &            91\,121 &           239\,786 &           84\,908 &   \textbf{71\,137} &   417\,294 \\
                          &  \sout{28} &           311\,375 &           252\,341 &  \textbf{98\,109} &           237\,270 &   912\,562 \\
                          &  \sout{29} &           144\,001 &           350\,208 &          151\,183 &  \textbf{132\,747} & 2\,561\,235 \\
                          &         30 &  \textbf{145\,468} &           464\,029 &          156\,648 &           285\,323 & 3\,058\,806 \\
                          &         31 &           381\,453 &  \textbf{130\,569} &          425\,412 &           264\,940 & 2\,508\,802 \\
 \hline\textbf{conflicts} &  \sout{25} &            21\,663 &            19\,473 &           21\,694 &   \textbf{18\,113} &   106\,940 \\
                          &  \sout{26} &   \textbf{21\,750} &            26\,160 &           24\,368 &            24\,112 &   154\,251 \\
                          &  \sout{27} &   \textbf{28\,000} &            46\,206 &           34\,613 &            29\,302 &   284\,788 \\
                          &  \sout{28} &            55\,972 &            48\,958 &  \textbf{41\,437} &            55\,795 &   670\,627 \\
                          &  \sout{29} &   \textbf{64\,615} &            76\,098 &           70\,879 &            65\,578 &   956\,430 \\
                          &         30 &            55\,678 &            92\,900 &           67\,725 &   \textbf{54\,496} &   753\,632 \\
                          &         31 &   \textbf{35\,980} &            46\,615 &           58\,403 &            38\,162 &   576\,474 \\
    \hline\textbf{rules} &  \sout{25} &  \textbf{151\,147} &           172\,162 &          167\,464 &           166\,327 &   128\,422 \\
                        &  \sout{26} &  \textbf{158\,522} &           180\,413 &          175\,517 &           174\,322 &   134\,873 \\
                        &  \sout{27} &  \textbf{165\,897} &           188\,664 &          183\,570 &           182\,317 &   141\,324 \\
                        &  \sout{28} &  \textbf{173\,272} &           196\,915 &          191\,623 &           190\,312 &   147\,775 \\
                        &  \sout{29} &  \textbf{180\,647} &           205\,166 &          199\,676 &           198\,307 &   154\,226 \\
                        &         30 &  \textbf{188\,022} &           213\,417 &          207\,729 &           206\,302 &   160\,677 \\
                        &         31 &  \textbf{195\,397} &           221\,668 &          215\,782 &           214\,297 &   167\,128 \\
      \hline\textbf{constraints} &  \sout{25} &           236\,285 &           234\,560 &          234\,524 &  \textbf{223\,456} &   206\,819 \\
                          &  \sout{26} &           249\,887 &           248\,306 &          248\,270 &  \textbf{236\,284} &   218\,864 \\
                          &  \sout{27} &           263\,489 &           262\,052 &          262\,016 &  \textbf{249\,112} &   230\,909 \\
                          &  \sout{28} &           277\,091 &           275\,798 &          275\,762 &  \textbf{261\,940} &   242\,954 \\
                          &  \sout{29} &           290\,693 &           289\,544 &          289\,508 &  \textbf{274\,768} &   254\,999 \\
                          &         30 &           304\,295 &           303\,290 &          303\,254 &  \textbf{287\,596} &   267\,044 \\
                          &         31 &           317\,897 &           317\,036 &          317\,000 &  \textbf{300\,424} &   279\,089 \\
\hline
\end{tabular}
\end{table}
 \begin{table}[h!]
\centering
\caption{Statistics for constraint $\varphi_2$ and the 2 robots instance.   }
\label{tbl:eval:d2:r2}
\begin{tabular}{|ll|rrrr|r|}
\hline
                          &  $\lambda$ &            $\WFA$ &          $\WFMm$ &          $\WFMs$ &           $\WFT$ &  $\WFNC$ \\

    \hline\textbf{translation time} &  \sout{25} &             2\,182 &           33\,091 &            4\,966 &   \textbf{2\,107} &     285 \\
                        &  \sout{26} &               529 &     \textbf{522} &              670 &            3\,006 &     444 \\
                        &         27 &      \textbf{548} &              904 &              624 &            2\,150 &     560 \\
                        &         28 &      \textbf{592} &              620 &              633 &            2\,372 &     377 \\
                        &         29 &               504 &              719 &     \textbf{478} &            2\,533 &     729 \\
                        &         30 &               653 &              488 &     \textbf{476} &            3\,254 &     807 \\
                        &         31 &      \textbf{650} &              714 &              711 &            3\,572 &     577 \\
    \hline\textbf{clingo time} &  \sout{25} &             5\,815 &   \textbf{5\,712} &            6\,921 &            7\,081 &   9\,845 \\
                        &  \sout{26} &             8\,918 &           10\,307 &           12\,229 &   \textbf{7\,465} &  13\,729 \\
                        &         27 &            13\,902 &            9\,206 &   \textbf{8\,105} &           11\,135 &  30\,361 \\
                        &         28 &            11\,543 &           11\,859 &            9\,095 &   \textbf{6\,658} &  28\,158 \\
                        &         29 &            10\,057 &   \textbf{7\,346} &           10\,258 &            7\,486 &  25\,309 \\
                        &         30 &            13\,708 &  \textbf{10\,204} &           10\,993 &           16\,829 &  23\,784 \\
                        &         31 &            15\,338 &           13\,696 &   \textbf{9\,805} &           12\,628 &  29\,339 \\
   \hline\textbf{choices} &  \sout{25} &   \textbf{33\,023} &           39\,073 &           42\,068 &           33\,878 &  51\,711 \\
                          &  \sout{26} &            45\,972 &           50\,647 &           62\,204 &  \textbf{39\,592} &  91\,708 \\
                          &         27 &            61\,491 &           55\,236 &  \textbf{50\,548} &           53\,343 & 120\,714 \\
                          &         28 &            59\,980 &           65\,021 &           57\,297 &  \textbf{38\,631} & 124\,386 \\
                          &         29 &            57\,594 &          188\,241 &           67\,603 &  \textbf{42\,736} & 121\,999 \\
                          &         30 &            71\,743 &  \textbf{66\,340} &           72\,332 &           80\,171 & 125\,925 \\
                          &         31 &            80\,675 &           75\,532 &          285\,296 &  \textbf{69\,654} & 150\,685 \\
 \hline\textbf{conflicts} &  \sout{25} &            18\,989 &           19\,514 &  \textbf{18\,501} &           19\,431 &  31\,833 \\
                          &  \sout{26} &            27\,180 &           27\,906 &           33\,399 &  \textbf{20\,912} &  40\,969 \\
                          &         27 &            38\,262 &           26\,728 &  \textbf{24\,532} &           30\,634 &  80\,625 \\
                          &         28 &            34\,453 &           33\,157 &           26\,112 &  \textbf{18\,682} &  77\,873 \\
                          &         29 &            31\,597 &  \textbf{18\,412} &           28\,473 &           20\,295 &  71\,170 \\
                          &         30 &            40\,803 &  \textbf{28\,115} &           30\,613 &           43\,978 &  71\,750 \\
                          &         31 &            44\,385 &           35\,627 &  \textbf{23\,957} &           35\,219 &  84\,896 \\
    \hline\textbf{rules} &  \sout{25} &   \textbf{73\,406} &          106\,969 &          111\,079 &           78\,734 &  67\,749 \\
                        &  \sout{26} &   \textbf{77\,106} &          112\,126 &          116\,414 &           82\,663 &  71\,213 \\
                        &         27 &   \textbf{80\,806} &          117\,283 &          121\,749 &           86\,592 &  74\,677 \\
                        &         28 &   \textbf{84\,506} &          122\,440 &          127\,084 &           90\,521 &  78\,141 \\
                        &         29 &   \textbf{88\,206} &          127\,597 &          132\,419 &           94\,450 &  81\,605 \\
                        &         30 &   \textbf{91\,906} &          132\,754 &          137\,754 &           98\,379 &  85\,069 \\
                        &         31 &   \textbf{95\,606} &          137\,911 &          143\,089 &          102\,308 &  88\,533 \\
      \hline\textbf{constraints} &  \sout{25} &  \textbf{118\,320} &          142\,056 &          146\,621 &          119\,776 & 113\,198 \\
                          &  \sout{26} &  \textbf{125\,251} &          150\,471 &          155\,306 &          126\,717 & 119\,809 \\
                          &         27 &  \textbf{132\,182} &          158\,886 &          163\,991 &          133\,658 & 126\,420 \\
                          &         28 &  \textbf{139\,113} &          167\,301 &          172\,676 &          140\,599 & 133\,031 \\
                          &         29 &  \textbf{146\,044} &          175\,716 &          181\,361 &          147\,540 & 139\,642 \\
                          &         30 &  \textbf{152\,975} &          184\,131 &          190\,046 &          154\,481 & 146\,253 \\
                          &         31 &  \textbf{159\,906} &          192\,546 &          198\,731 &          161\,422 & 152\,864 \\
\hline
\end{tabular}
\end{table}
 \begin{table}[h!]
\centering
\caption{Statistics for constraint $\varphi_2$ and the 3 robots instance.   }
\label{tbl:eval:d2:r3}
\begin{tabular}{|ll|rrrr|r|}
\hline
                          &  $\lambda$ &            $\WFA$ &           $\WFMm$ &       $\WFMs$ &            $\WFT$ &    $\WFNC$ \\

    \hline\textbf{translation time} &  \sout{25} &    \textbf{1\,632} &            45\,303 &         4\,973 &             2\,718 &       317 \\
                        &  \sout{26} &               567 &      \textbf{479} &           718 &             3\,389 &       450 \\
                        &  \sout{27} &      \textbf{557} &               889 &           635 &             2\,676 &       533 \\
                        &  \sout{28} &               599 &               652 &  \textbf{558} &             2\,779 &       523 \\
                        &  \sout{29} &               564 &               718 &  \textbf{490} &             3\,160 &       800 \\
                        &  \sout{30} &               680 &               495 &  \textbf{455} &             3\,289 &       888 \\
                        &         31 &      \textbf{683} &               762 &           769 &             3\,786 &       592 \\
    \hline\textbf{clingo time} &  \sout{25} &            17\,910 &   \textbf{14\,564} &        16\,993 &            14\,647 &    47\,761 \\
                        &  \sout{26} &            33\,219 &   \textbf{26\,979} &        36\,179 &            29\,358 &    75\,161 \\
                        &  \sout{27} &            40\,397 &            42\,075 &        50\,820 &   \textbf{35\,071} &   195\,261 \\
                        &  \sout{28} &            74\,244 &   \textbf{59\,909} &        78\,488 &            69\,510 &   539\,328 \\
                        &  \sout{29} &           110\,278 &           104\,918 &       122\,672 &   \textbf{91\,111} &   786\,940 \\
                        &  \sout{30} &  \textbf{177\,537} &           197\,182 &       402\,643 &           186\,005 &   496\,808 \\
                        &         31 &           252\,945 &  \textbf{117\,736} &       234\,057 &           172\,171 &   376\,699 \\
   \hline\textbf{choices} &  \sout{25} &            79\,355 &            89\,839 &        82\,854 &   \textbf{68\,478} &   173\,746 \\
                          &  \sout{26} &           126\,138 &           117\,782 &       137\,992 &  \textbf{113\,770} &   245\,978 \\
                          &  \sout{27} &           141\,902 &           167\,256 &       178\,598 &  \textbf{131\,983} &   417\,294 \\
                          &  \sout{28} &           222\,615 &           215\,795 &       215\,028 &  \textbf{199\,998} &   912\,562 \\
                          &  \sout{29} &           284\,038 &           300\,505 &       279\,271 &  \textbf{251\,921} & 2\,561\,235 \\
                          &  \sout{30} &  \textbf{388\,048} &           464\,021 &     1\,733\,075 &           401\,158 & 3\,058\,806 \\
                          &         31 &           539\,032 &  \textbf{348\,336} &     1\,475\,025 &           430\,008 & 2\,508\,802 \\
 \hline\textbf{conflicts} &  \sout{25} &            41\,110 &            39\,086 &        37\,373 &   \textbf{33\,142} &   106\,940 \\
                          &  \sout{26} &            72\,849 &   \textbf{58\,218} &        69\,422 &            61\,248 &   154\,251 \\
                          &  \sout{27} &            81\,352 &            86\,506 &        95\,104 &   \textbf{71\,633} &   284\,788 \\
                          &  \sout{28} &           137\,837 &           117\,090 &       120\,923 &  \textbf{117\,023} &   670\,627 \\
                          &  \sout{29} &           181\,467 &           180\,692 &       163\,538 &  \textbf{151\,996} &   956\,430 \\
                          &  \sout{30} &           259\,733 &           294\,888 &       434\,690 &  \textbf{258\,368} &   753\,632 \\
                          &         31 &           364\,780 &  \textbf{203\,024} &       302\,537 &           260\,559 &   576\,474 \\
    \hline\textbf{rules} &  \sout{25} &  \textbf{137\,614} &           188\,131 &       201\,109 &           145\,489 &   128\,422 \\
                        &  \sout{26} &  \textbf{144\,449} &           197\,159 &       210\,695 &           152\,663 &   134\,873 \\
                        &  \sout{27} &  \textbf{151\,284} &           206\,187 &       220\,281 &           159\,837 &   141\,324 \\
                        &  \sout{28} &  \textbf{158\,119} &           215\,215 &       229\,867 &           167\,011 &   147\,775 \\
                        &  \sout{29} &  \textbf{164\,954} &           224\,243 &       239\,453 &           174\,185 &   154\,226 \\
                        &  \sout{30} &  \textbf{171\,789} &           233\,271 &       249\,039 &           181\,359 &   160\,677 \\
                        &         31 &  \textbf{178\,624} &           242\,299 &       258\,625 &           188\,533 &   167\,128 \\
      \hline\textbf{constraints} &  \sout{25} &  \textbf{215\,577} &           253\,048 &       263\,456 &           217\,522 &   206\,819 \\
                          &  \sout{26} &  \textbf{228\,144} &           267\,841 &       278\,888 &           230\,104 &   218\,864 \\
                          &  \sout{27} &  \textbf{240\,711} &           282\,634 &       294\,320 &           242\,686 &   230\,909 \\
                          &  \sout{28} &  \textbf{253\,278} &           297\,427 &       309\,752 &           255\,268 &   242\,954 \\
                          &  \sout{29} &  \textbf{265\,845} &           312\,220 &       325\,184 &           267\,850 &   254\,999 \\
                          &  \sout{30} &  \textbf{278\,412} &           327\,013 &       340\,616 &           280\,432 &   267\,044 \\
                          &         31 &  \textbf{290\,979} &           341\,806 &       356\,048 &           293\,014 &   279\,089 \\
\hline
\end{tabular}
\end{table}
 \begin{table}[h!]
\centering
\caption{Statistics for constraint $\varphi_3$ and the 2 robots instance.   }
\label{tbl:eval:d3:r2}
\begin{tabular}{|ll|rrrr|r|}
\hline
                          &  $\lambda$ &            $\WFA$ &         $\WFMm$ &         $\WFMs$ &            $\WFT$ &  $\WFNC$ \\

    \hline\textbf{translation time} &  \sout{25} &    \textbf{2\,533} &  \color{red}{-} &          12\,682 &             3\,343 &     260 \\
                        &  \sout{26} &      \textbf{517} &  \color{red}{-} &             926 &             3\,717 &     436 \\
                        &  \sout{27} &               606 &  \color{red}{-} &    \textbf{578} &             2\,781 &     599 \\
                        &  \sout{28} &      \textbf{623} &  \color{red}{-} &             834 &             2\,864 &     381 \\
                        &  \sout{29} &      \textbf{547} &  \color{red}{-} &             739 &             3\,429 &     761 \\
                        &         30 &      \textbf{654} &  \color{red}{-} &             754 &             3\,276 &     822 \\
                        &         31 &      \textbf{652} &  \color{red}{-} &  \color{red}{-} &             3\,613 &     630 \\
    \hline\textbf{clingo time} &  \sout{25} &            12\,253 &  \color{red}{-} &         229\,981 &    \textbf{9\,291} &   9\,904 \\
                        &  \sout{26} &            25\,188 &  \color{red}{-} &         359\,347 &   \textbf{13\,359} &  14\,406 \\
                        &  \sout{27} &            41\,912 &  \color{red}{-} &         575\,544 &   \textbf{12\,090} &  30\,541 \\
                        &  \sout{28} &            42\,680 &  \color{red}{-} &         898\,255 &   \textbf{16\,686} &  28\,792 \\
                        &  \sout{29} &            46\,213 &  \color{red}{-} &         837\,573 &   \textbf{18\,315} &  24\,138 \\
                        &         30 &            36\,066 &  \color{red}{-} &         706\,322 &   \textbf{17\,020} &  23\,975 \\
                        &         31 &            37\,229 &  \color{red}{-} &  \color{red}{-} &   \textbf{36\,629} &  29\,704 \\
   \hline\textbf{choices} &  \sout{25} &            54\,885 &  \color{red}{-} &      15\,989\,429 &   \textbf{44\,122} &  51\,711 \\
                          &  \sout{26} &            96\,641 &  \color{red}{-} &      26\,420\,540 &   \textbf{57\,103} &  91\,708 \\
                          &  \sout{27} &           140\,631 &  \color{red}{-} &      53\,728\,668 &   \textbf{53\,232} & 120\,714 \\
                          &  \sout{28} &           140\,053 &  \color{red}{-} &      75\,682\,776 &   \textbf{70\,231} & 124\,386 \\
                          &  \sout{29} &           148\,264 &  \color{red}{-} &      70\,859\,664 &   \textbf{78\,123} & 121\,999 \\
                          &         30 &           130\,536 &  \color{red}{-} &      29\,387\,312 &   \textbf{71\,563} & 125\,925 \\
                          &         31 &           133\,751 &  \color{red}{-} &  \color{red}{-} &  \textbf{131\,548} & 150\,685 \\
 \hline\textbf{conflicts} &  \sout{25} &            38\,229 &  \color{red}{-} &          42\,422 &   \textbf{28\,841} &  31\,833 \\
                          &  \sout{26} &            70\,619 &  \color{red}{-} &          54\,901 &   \textbf{38\,380} &  40\,969 \\
                          &  \sout{27} &           106\,770 &  \color{red}{-} &          91\,584 &   \textbf{35\,137} &  80\,625 \\
                          &  \sout{28} &           105\,411 &  \color{red}{-} &         123\,025 &   \textbf{47\,733} &  77\,873 \\
                          &  \sout{29} &           110\,880 &  \color{red}{-} &         120\,513 &   \textbf{52\,747} &  71\,170 \\
                          &         30 &            95\,068 &  \color{red}{-} &         106\,837 &   \textbf{46\,265} &  71\,750 \\
                          &         31 &            96\,986 &  \color{red}{-} &  \color{red}{-} &   \textbf{93\,448} &  84\,896 \\
        \hline\textbf{rules} &  \sout{25} &   \textbf{82\,732} &  \color{red}{-} &       3\,970\,094 &            89\,873 &  67\,749 \\
                            &  \sout{26} &   \textbf{86\,824} &  \color{red}{-} &       4\,151\,638 &            94\,270 &  71\,213 \\
                            &  \sout{27} &   \textbf{90\,916} &  \color{red}{-} &       4\,333\,182 &            98\,667 &  74\,677 \\
                            &  \sout{28} &   \textbf{95\,008} &  \color{red}{-} &       4\,514\,726 &           103\,064 &  78\,141 \\
                            &  \sout{29} &   \textbf{99\,100} &  \color{red}{-} &       4\,696\,270 &           107\,461 &  81\,605 \\
                            &         30 &  \textbf{103\,192} &  \color{red}{-} &       4\,877\,814 &           111\,858 &  85\,069 \\
                            &         31 &  \textbf{107\,284} &  \color{red}{-} &  \color{red}{-} &           116\,255 &  88\,533 \\
      \hline\textbf{constraints} &  \sout{25} &           130\,874 &  \color{red}{-} &       2\,856\,055 &  \textbf{126\,120} & 113\,198 \\
                          &  \sout{26} &           138\,567 &  \color{red}{-} &       3\,057\,106 &  \textbf{133\,359} & 119\,809 \\
                          &  \sout{27} &           146\,260 &  \color{red}{-} &       3\,258\,157 &  \textbf{140\,598} & 126\,420 \\
                          &  \sout{28} &           153\,953 &  \color{red}{-} &       3\,459\,208 &  \textbf{147\,837} & 133\,031 \\
                          &  \sout{29} &           161\,646 &  \color{red}{-} &       3\,660\,259 &  \textbf{155\,076} & 139\,642 \\
                          &         30 &           169\,339 &  \color{red}{-} &       3\,861\,310 &  \textbf{162\,315} & 146\,253 \\
                          &         31 &           177\,032 &  \color{red}{-} &  \color{red}{-} &  \textbf{169\,554} & 152\,864 \\
\hline
\end{tabular}
\end{table}
 \begin{table}[h!]
\centering
\caption{Statistics for constraint $\varphi_3$ and the 3 robots instance.   }
\label{tbl:eval:d3:r3}
\begin{tabular}{|ll|rrrr|r|}
\hline
                          &  $\lambda$ &            $\WFA$ &         $\WFMm$ &         $\WFMs$ &            $\WFT$ &    $\WFNC$ \\

    \hline\textbf{translation time} &  \sout{25} &    \textbf{3\,112} &  \color{red}{-} &          11\,001 &             3\,278 &       271 \\
                        &  \sout{26} &      \textbf{481} &  \color{red}{-} &             864 &             3\,895 &       441 \\
                        &  \sout{27} &      \textbf{621} &  \color{red}{-} &             851 &             3\,314 &       546 \\
                        &  \sout{28} &      \textbf{627} &  \color{red}{-} &             760 &             3\,375 &       449 \\
                        &  \sout{29} &      \textbf{531} &  \color{red}{-} &             708 &             3\,855 &       777 \\
                        &  \sout{30} &      \textbf{679} &  \color{red}{-} &  \color{red}{-} &             3\,351 &       815 \\
                        &  \sout{31} &      \textbf{663} &  \color{red}{-} &  \color{red}{-} &             4\,519 &       555 \\
    \hline\textbf{clingo time} &  \sout{25} &            23\,083 &  \color{red}{-} &         407\,903 &   \textbf{14\,376} &    49\,111 \\
                        &  \sout{26} &            29\,884 &  \color{red}{-} &         364\,093 &   \textbf{21\,753} &    74\,915 \\
                        &  \sout{27} &            36\,847 &  \color{red}{-} &         472\,354 &   \textbf{30\,569} &   195\,389 \\
                        &  \sout{28} &            60\,786 &  \color{red}{-} &         476\,299 &   \textbf{45\,751} &   526\,867 \\
                        &  \sout{29} &           107\,503 &  \color{red}{-} &       1\,105\,621 &   \textbf{78\,207} &   796\,914 \\
                        &  \sout{30} &           151\,375 &  \color{red}{-} &  \color{red}{-} &  \textbf{108\,197} &   500\,901 \\
                        &  \sout{31} &           248\,384 &  \color{red}{-} &  \color{red}{-} &  \textbf{163\,082} &   378\,597 \\
   \hline\textbf{choices} &  \sout{25} &            98\,424 &  \color{red}{-} &      36\,202\,304 &   \textbf{60\,086} &   173\,746 \\
                          &  \sout{26} &           110\,269 &  \color{red}{-} &      26\,575\,984 &   \textbf{86\,056} &   245\,978 \\
                          &  \sout{27} &           111\,872 &  \color{red}{-} &      32\,814\,318 &  \textbf{103\,601} &   417\,294 \\
                          &  \sout{28} &           170\,755 &  \color{red}{-} &         297\,764 &  \textbf{142\,450} &   912\,562 \\
                          &  \sout{29} &           259\,262 &  \color{red}{-} &      57\,960\,092 &  \textbf{200\,553} & 2\,561\,235 \\
                          &  \sout{30} &           310\,797 &  \color{red}{-} &  \color{red}{-} &  \textbf{262\,182} & 3\,058\,806 \\
                          &  \sout{31} &           460\,825 &  \color{red}{-} &  \color{red}{-} &  \textbf{350\,260} & 2\,508\,802 \\
 \hline\textbf{conflicts} &  \sout{25} &            63\,019 &  \color{red}{-} &          97\,082 &   \textbf{36\,166} &   106\,940 \\
                          &  \sout{26} &            72\,644 &  \color{red}{-} &          81\,166 &   \textbf{53\,689} &   154\,251 \\
                          &  \sout{27} &            75\,839 &  \color{red}{-} &         129\,245 &   \textbf{66\,361} &   284\,788 \\
                          &  \sout{28} &           116\,876 &  \color{red}{-} &         136\,306 &   \textbf{93\,099} &   670\,627 \\
                          &  \sout{29} &           184\,608 &  \color{red}{-} &         249\,421 &  \textbf{137\,604} &   956\,430 \\
                          &  \sout{30} &           230\,970 &  \color{red}{-} &  \color{red}{-} &  \textbf{185\,500} &   753\,632 \\
                          &  \sout{31} &           347\,612 &  \color{red}{-} &  \color{red}{-} &  \textbf{254\,287} &   576\,474 \\
    \hline\textbf{rules} &  \sout{25} &  \textbf{151\,581} &  \color{red}{-} &       3\,083\,856 &           162\,198 &   128\,422 \\
                        &  \sout{26} &  \textbf{159\,003} &  \color{red}{-} &       3\,225\,088 &           170\,074 &   134\,873 \\
                        &  \sout{27} &  \textbf{166\,425} &  \color{red}{-} &       3\,366\,320 &           177\,950 &   141\,324 \\
                        &  \sout{28} &  \textbf{173\,847} &  \color{red}{-} &       3\,507\,552 &           185\,826 &   147\,775 \\
                        &  \sout{29} &  \textbf{181\,269} &  \color{red}{-} &       3\,648\,784 &           193\,702 &   154\,226 \\
                        &  \sout{30} &  \textbf{188\,691} &  \color{red}{-} &  \color{red}{-} &           201\,578 &   160\,677 \\
                        &  \sout{31} &  \textbf{196\,113} &  \color{red}{-} &  \color{red}{-} &           209\,454 &   167\,128 \\
      \hline\textbf{constraints} &  \sout{25} &           235\,091 &  \color{red}{-} &       2\,435\,143 &  \textbf{227\,044} &   206\,819 \\
                          &  \sout{26} &           248\,801 &  \color{red}{-} &       2\,594\,044 &  \textbf{240\,073} &   218\,864 \\
                          &  \sout{27} &           262\,511 &  \color{red}{-} &       2\,752\,945 &  \textbf{253\,102} &   230\,909 \\
                          &  \sout{28} &           276\,221 &  \color{red}{-} &       2\,911\,846 &  \textbf{266\,131} &   242\,954 \\
                          &  \sout{29} &           289\,931 &  \color{red}{-} &       3\,070\,747 &  \textbf{279\,160} &   254\,999 \\
                          &  \sout{30} &           303\,641 &  \color{red}{-} &  \color{red}{-} &  \textbf{292\,189} &   267\,044 \\
                          &  \sout{31} &           317\,351 &  \color{red}{-} &  \color{red}{-} &  \textbf{305\,218} &   279\,089 \\
\hline
\end{tabular}
\end{table}
  
\end{document}